\documentclass{article} 
\usepackage{iclr2027_conference,times}

\usepackage{subcaption}

\usepackage{amsmath,amsfonts,bm}

\def\eqref#1{equation~\ref{#1}}

\def\1{\bm{1}}

\DeclareMathAlphabet{\mathsfit}{\encodingdefault}{\sfdefault}{m}{sl}
\SetMathAlphabet{\mathsfit}{bold}{\encodingdefault}{\sfdefault}{bx}{n}

\usepackage[utf8]{inputenc} 
\usepackage[T1]{fontenc}    
\usepackage{hyperref}       
\usepackage{url}            
\usepackage{booktabs}       
\usepackage{amsfonts}       
\usepackage{nicefrac}       
\usepackage{microtype}      
\usepackage{xcolor}         

\usepackage{hyperref}
\usepackage{url}
\usepackage{tcolorbox}
\usepackage{xcolor}
\usepackage{amsmath}
\usepackage{arydshln}
\usepackage{wrapfig}
\usepackage{graphicx}
\usepackage{placeins}

\newcommand{\datasetname}{Sci-$\rho$~}
\newcommand{\datasetsize}{606}

\title{
Sci-$\rho$: A Multilingual Visually-Grounded Symbolic Benchmark for STEM Problems
}

\author{
Muhammad Falensi Azmi$^{\dagger *}$ \quad Ikhlasul Akmal Hanif$^{\dagger *}$ \quad Vallerie Alexandra Putra$^{\ddagger}$ \\
\textbf{Adi Yeltay}$^{\dagger}$ \quad \textbf{Abdullah Mubarak}\textsuperscript{\S} \quad \textbf{Fajri Koto}$^{\dagger}$ \\
\normalfont $^{\dagger}$MBZUAI \quad $^{\ddagger}$Binus University \quad \textsuperscript{\S}Bandung Institute of Technology \\
\normalfont \texttt{\{muhammad.azmi, ikhlasul.hanif\}@mbzuai.ac.ae} \\
\normalfont\small $^{*}$Equal contribution
}

\iclrfinalcopy 
\begin{document}

\maketitle

\begin{abstract}

Symbolic benchmarks have emerged as a key approach to assess model robustness under minor modifications to STEM-related questions. However, existing symbolic benchmarks mostly remain limited to mathematical reasoning, lack visual grounding, and are predominantly in English. In this work, we introduce Sci-$\rho$ (\textbf{Sci}ence \textbf{Rho}bustness), a dynamic benchmark for visually-grounded STEM problems spanning five subjects and seven languages, comprising 4,242 problem templates (606 per language) crafted by domain experts, including Olympiad medalists. Each template is implemented as \textit{executable Python code} that generates diverse but equivalent problem instances by varying numerical values, visual patterns, geometric shapes, color schemes, and function types, resulting in 42,420 instances in total, each paired with reasoning steps and ground-truth solutions. We evaluated 17 state-of-the-art VLMs and discovered a noticeable gap between worst-case accuracy (defined as the proportion of problem templates that a model answers correctly across every generated variation) and average accuracy. We also discovered that smaller models show noticeable performance degradation across languages, whereas proprietary and larger models remain robust. Step-level evaluation reflects this same trend, revealing a significant gap between average F1 and worst-case F1 scores. Finally, our inspection of attention heads of a VLM reveals substantial cross-lingual variation in the relative attention allocated to image tokens compared to text tokens. Our work highlights the importance of evaluation beyond static benchmarks as a metric to measure the quality of VLMs.

\end{abstract}

\section{Introduction} \label{sec:introduction}

Large Vision-Language Models (VLMs) have demonstrated notable capabilities in solving visually-grounded tasks across various domains~\citep{ren2025vgrpbenchvisualgridreasoning, shinoda2025agrobenchvisionlanguagemodelbenchmark, nayak-etal-2024-benchmarking, nyandwi-etal-2025-grounding}. To measure progress on reasoning that requires visual understanding, the community has introduced a wide range of benchmarks: \texttt{MATH-V}~\citep{wang2024measuringmultimodalmathematicalreasoning}, \texttt{MathVista}~\citep{lu2024mathvistaevaluatingmathematicalreasoning}, and \texttt{MM-Math}~\citep{sun-etal-2024-mm} for Mathematics; \texttt{SeePhys}~\citep{xiang2025seephysdoesseeinghelp}, \texttt{PHYBench}~\citep{qiu2025phybenchholisticevaluationphysical}, and \texttt{PhysReason}~\citep{zhang2025physreasoncomprehensivebenchmarkphysicsbased} for Physics; and \texttt{SCIENCEQA}~\citep{lu2022learnexplainmultimodalreasoning} and \texttt{SciVerse}~\citep{guo2025sciverseunveilingknowledgecomprehension} for more general science questions. Across these benchmarks, VLMs consistently fall short of human performance.

Although these benchmarks represent an important early effort, the robustness of VLM problem-solving remains largely unexplored. Existing evaluations mostly rely on static, one-off images: each problem is rendered exactly once, and the model is scored on that single instance. Yet a problem's visual presentation is incidental to its underlying logic--the same geometric relationship, for instance, can be drawn with different side lengths without altering the reasoning required to solve it. Humans solve such variants without difficulty; whether VLMs do is an open question. Without controlled perturbations of the image itself, it is impossible to distinguish models that genuinely reason about visual content from those that succeed by matching surface features of problems seen during training.



\begin{figure}[t]
    \centering
    \includegraphics[width=1.0\textwidth]{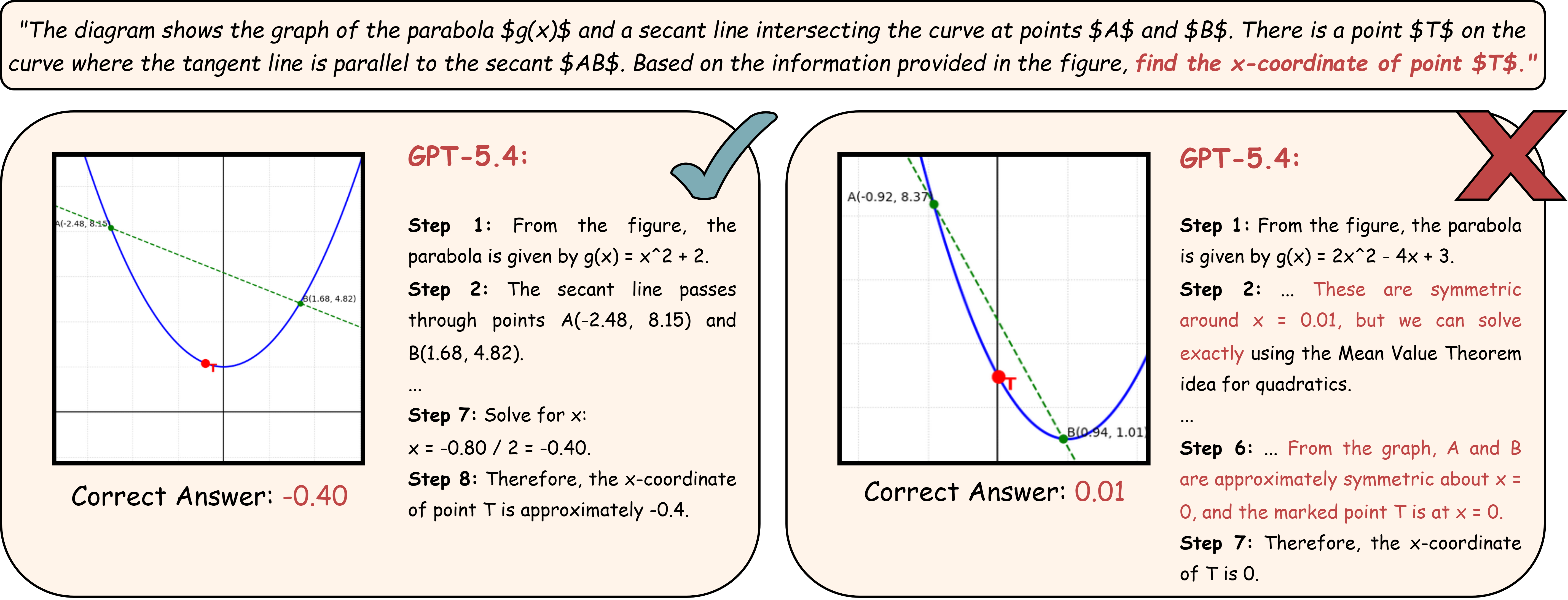}
    \caption{An illustrative failure case of GPT-5.4 when evaluated across problem variants in our dataset, highlighting vulnerabilities in VLM robustness.}
    \label{fig:sample_fail}
\end{figure}

Figure~\ref{fig:sample_fail} shows an example where GPT-5.4---a frontier model---successfully answers a problem in one scenario but fails in another seemingly similar question where the graph and point coordinates are varied. This inconsistency suggests a lack of robust reasoning capability and highlights the importance of evaluation beyond static benchmarks.

A further limitation of prior work is its narrow scope. Coverage is concentrated on Mathematics and Physics, leaving the rest of STEM--Biology, Chemistry, and Computer Science--largely outside the scope of visually-grounded evaluation, despite each presenting distinct visual reasoning demands such as molecular structures, biological diagrams, and algorithmic traces. Evaluation is also conducted almost entirely in English, leaving open whether observed capabilities transfer to the multilingual settings in which most learners and practitioners actually encounter STEM content.

We address these gaps by introducing the first symbolic, multilingual robustness benchmark for visually-grounded STEM reasoning. Our benchmark spans Mathematics, Physics, Biology, Chemistry, and Computer Science at the high school and pre-university level, with each problem implemented as a Python template that dynamically generates question variants—along with their reasoning steps and solutions—across seven languages. These templates were carefully constructed and reviewed by domain experts, including science olympiad medalists, degree holders, and final-year undergraduates with demonstrated expertise in their respective fields. Our contributions are as follows:

\begin{itemize}
\item We developed \datasetname{} (\textbf{Sci}ence \textbf{Rho}bustness) as a multilingual multimodal symbolic benchmark to assess VLMs capabilities in solving STEM-Problems. The dataset consists of $42,420$ problem instances generated from \datasetsize{} templates, 7 languages, and 10 variants in five core STEM subjects: Mathematics, Physics, Chemistry, Biology, and Computer Science.
\item We benchmarked 17 VLMs using \datasetname{}. Our evaluation provides an in-depth analysis of the performance gap between worst-case and average accuracy, the cross-lingual performance degradation relative to English, and the step-level validity of the models' reasoning compared to ground-truth solutions.
\item We perform a mechanistic analysis of how VLMs allocate attention between visual and textual content during STEM reasoning. We define the Image Attention Ratio (IAR) to quantify the magnitude of attention paid to image tokens relative to textual tokens, and compare IAR across languages and subjects on a representative open-weight VLM.


\end{itemize}

\section{Related Works} \label{sec:related_works}

\paragraph{Visual Language Models}

Vision-Language Models (VLMs) have advanced at a remarkable pace, with proprietary models like GPT-5~\citep{singh2025openaigpt5card} and Gemini 2.5~\citep{deepmind2025gemini25pro} setting the state of the art. The open-source community has kept pace through diverse architectural innovations: Gemma 3's optimized local-to-global attention ratios~\citep{gemmateam2025gemma3technicalreport}, Qwen3's multi-level ViT feature integration, and the native multimodal capabilities of Qwen 3.5. Together with models such as Llama 4 Scout~\citep{meta_llama4_2025}, InternVL 3.5~\citep{wang2025internvl35advancingopensourcemultimodal}, SEA-LION VL~\citep{aisingapore2026sealionv4}, and Molmo 2~\citep{clark2026molmo2openweightsdata}, these systems continue to redefine performance on visual benchmarks. Yet robustness remains a critical concern~\citep{vo2026visionlanguagemodelsbiased, Irawan_2026_CVPR}: persistent failures on simple tasks such as object counting raise questions about the sufficiency of static benchmarks, particularly for rigorous STEM applications.

\paragraph{Robustness \& Symbolic Benchmark}
Considerable research effort has been directed toward evaluating the robustness of LLMs and VLMs. \cite{zheng2024largelanguagemodelsrobust} found that LLMs are vulnerable to changes in option position when solving Multiple Choice Questions due to their inherent selection bias. Another work by \cite{zong2024foolvisionandlanguage} discovered a similar option-shuffling vulnerability in VLMs. Recently, \cite{mirzadeh2025gsmsymbolicunderstandinglimitationsmathematical} discovered that LLMs show a non-negligible variance when solving the same textual mathematical question with different numerical values, indicating that current LLMs are not capable of genuine logical reasoning. Extensive study on VLMs by \cite{zou2025dynamathdynamicvisualbenchmark} concluded that VLMs are sensitive to changes in input variations when solving math problems, despite the knowledge and steps required to solve the problem remaining the same. 

Yet across this body of work, robustness evaluation through symbolic, programmatically generated variants for visually-grounded reasoning has been confined almost exclusively to mathematics and to English. To our knowledge, no prior benchmark provides symbolic, image-level perturbations for visually-grounded questions in Physics, Chemistry, Biology, or Computer Science, nor in any non-English language. Our work is the first to address both gaps in a single benchmark.


\begin{figure}[t]
    \centering
    \includegraphics[width=1.0\textwidth]{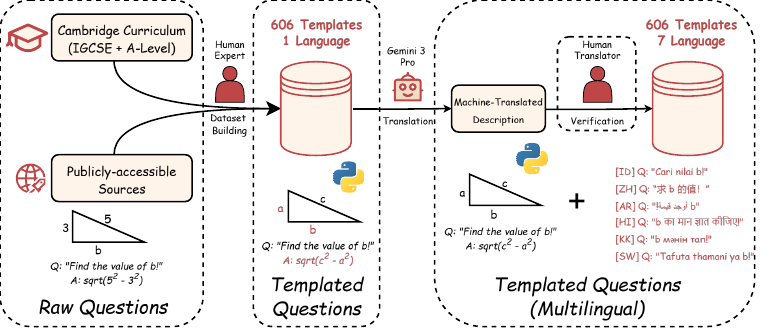}
    \caption{Dataset construction pipeline.} 
    \label{fig:pipeline}
    \vspace{-0.4cm}
\end{figure}

\section{\datasetname{} Dataset} \label{sec:dataset}

\subsection{Dataset Construction} \label{sec:dataset_construction}
Figure~\ref{fig:pipeline} illustrates the dataset construction pipeline, which was conducted in two stages: template creation and translation. In the first stage, we collected vision-dependent STEM problems from open sources and transformed them into dynamic Python classes, guided by Cambridge curriculum. In the second stage, we translated the problems into six additional languages with verification by native speakers.

\paragraph{Template Creation} We curated a list of vision-dependent materials based on the Cambridge curriculum as our ground reference\footnote{We use the Cambridge curriculum as a high-level guideline to structure topics and subtopics. However, we do not strictly follow this curriculum, and the dataset includes additional topics and problem types beyond it.}. This includes several topics from the Cambridge IGCSE and A-level curricula, spanning five subjects: Mathematics, Physics, Chemistry, Biology, and Computer Science. After identifying the relevant topics and subtopics to be included in our dataset, we collected problems from various publicly available sources, mostly from Cambridge past paper exams (IGCSE \& A-Level), previous benchmark, and originally-constructed questions, and subsequently assigned each problem to the appropriate topic and subtopic defined earlier. We only included problems that depends on image to solve, as not all problems in STEM require visual component.

We hired STEM graduates, advanced undergraduates, and science olympiad medalists as expert workers, and tasked them with manually rewriting each question into a Python class that serves as a dynamic template. For each template, we included several attributes critical to the problem (such as description, image, final answer) along with common metadata (topic, subtopic, source, curriculum, and question type). All calculations leading to the final answer were also computed and stored in each Python class. To provide a clear trace of the reasoning process, our dataset includes dynamically generated step-by-step solutions written in LaTeX. While there is no theoretical upper limit to the number of variants a template can generate, each is guaranteed to yield at least ten distinct problem instances. 

Finally, we applied strict quality control. Problem setters first cross-checked each other's work within the same subject, after which a dedicated team of quality control workers manually verified the outputs generated from each Python template, examining both the intermediate reasoning steps and the final answer for correctness. Further details are provided in Appendix~\ref{appendix:a1}. This stage yielded \datasetsize{} high-quality multimodal problem templates in English.

\paragraph{Translation} In this step, the dataset was expanded into six additional languages: Indonesian, Chinese, Arabic, Hindi, Kazakh, and Swahili. We first translated the descriptions for each template using Gemini 3 Pro while keeping the images in English. These machine-translated descriptions were then verified by native speakers of each corresponding language, involving both the authors and hired contributors. We also employed quality control by intentionally introducing errors into 40 of the \datasetsize{} descriptions (approximately 7\%) at random. Translators who failed to identify and correct these errors were required to revise their work (more details are provided in Appendix~\ref{appendix:a2}). Finally, we integrated these translations into our main dataset. The result is \datasetsize{} high-quality problem templates in seven languages with localized descriptions and a unified English visual component.

\begin{figure}[t]
    \centering
    \includegraphics[width=1.0\textwidth]{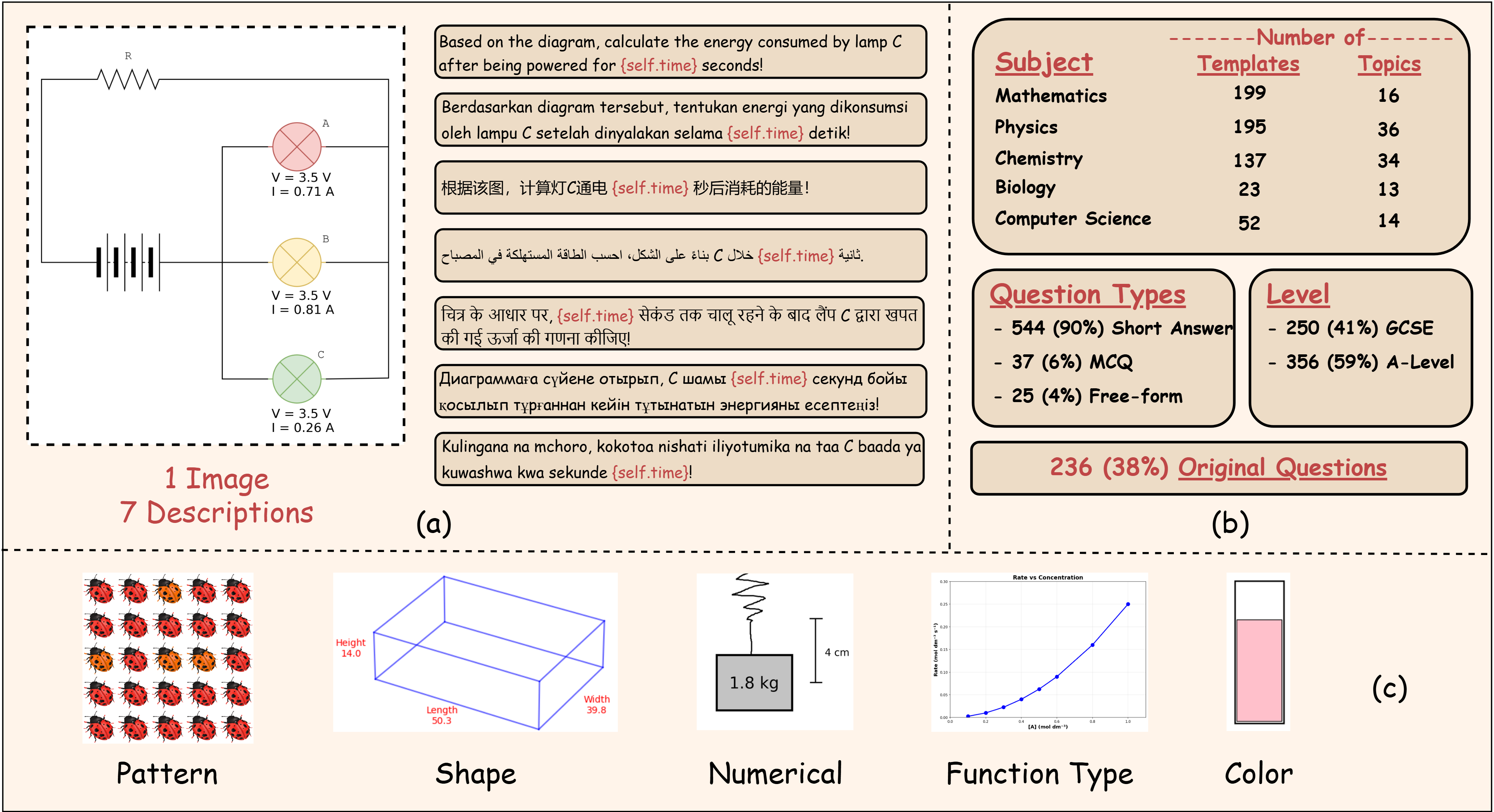}
    \caption{A glimpse of our dataset. (a) An example from our dataset where each template consists of one image and description in seven languages; (b) Statistics of our dataset; (c) Example of variants included in our dataset.}
    \label{fig:dataset_statistics}
\end{figure}

\subsection{Dataset Overview} \label{sec:dataset_overview}

Our dataset consists of \datasetsize{} high-quality templates written in Python, encompassing five STEM subjects (Mathematics, Physics, Chemistry, Biology, and Computer Science) across seven language variants (English, Indonesian, Chinese, Arabic, Hindi, Kazakh, and Swahili). Each template is categorized into exactly one topic and one subtopic, taking the form of a (numerical) Short Answer, MCQ, or Free-form problem, and incorporates several dimensions of variability.

Figure~\ref{fig:dataset_statistics} (part a) shows a sample taken from the Physics subset of our dataset, which consists of descriptions in seven languages and one unified image. Specifically for this template, both the information shown in the image, such as voltage (V) and current (I), and the variable \textit{self.time} in the description are dynamic. This setup allows us to evaluate performance variations when a VLM is prompted with diverse numerical information across different languages.

The statistics of our dataset are presented in Figure~\ref{fig:dataset_statistics} (part b). In general, our dataset is predominantly composed of Mathematics, Physics, and Chemistry problems, comprising 199, 195, and 137 templates respectively, followed by a smaller number of Computer Science (52) and Biology (23) templates. The number of topics for each subject is given in the figure. The majority of the dataset consists of Short Answer problems (accounting for 90\% of the total), followed by a smaller proportion of MCQs (6\%) and Free-form questions (4\%). Our dataset spans two difficulty levels, with more than half (59\%) of the problems being at the A-Level and 41\% at the GCSE level. About 38\% of the problems are original (designed from scratch), while the remainder are adapted from various open sources.

We identify five dimensions of variability in our dataset, as illustrated in Figure~\ref{fig:dataset_statistics} (part c). The \textit{pattern} and \textit{color} variants test the model capabilities to interpret positional variants (such as the arrangement of ladybugs) and diverse color schemes (such as the color of a chemical solution). To assess spatial and structural comprehension, the \textit{shape} variant introduces different geometric forms, such as varying cuboid dimensions. Furthermore, to probe quantitative and analytical capabilities, the \textit{numerical} variant challenges models by varying values embedded within either the text or the accompanying image, while the \textit{function type} variant measures proficiency in distinguishing mathematical functions (e.g., linear vs quadratic). More examples are provided in Appendix~\ref{appendix:b}.

\section{Experiment} \label{sec:experiment}
We evaluated Vision Language Models using six metrics: Average-case Accuracy ($\mathcal{A}_{avg}$), Worst-case Accuracy ($\mathcal{A}_{wst}$), Step Recall ($\mathcal{R}$), Step Precision ($\mathcal{P}$), Step F1 ($\mathcal{F}$), and Worst-case Step F1 ($\mathcal{F}_{wst}$), defined in Section~\ref{sec:evaluation_setup}. Additionally, we analyzed the attention heads of \textit{Qwen3-VL-8B-Instruct} to better understand how strongly the model attends to image tokens compared to text tokens.

\subsection{Evaluation Setup} \label{sec:evaluation_setup}


Our dataset consists of \datasetsize{} problem templates. From each template, we generated $M=10$ distinct problem instances across 7 languages, yielding a total of $10 \times \datasetsize{} \times 7 = 42,420$ unique problem instances. This comprehensive collection was then utilized to rigorously evaluate VLM robustness.

For our experiments, we benchmarked leading closed-weight models, including Gemini 2.5 Pro~\citep{deepmind2025gemini25pro}, Gemini 2.5 Flash~\citep{deepmind2025gemini25flash}, and GPT-5.4~\citep{singh2025openaigpt5card}, alongside open-weight models such as Llama 4 Scout~\citep{meta_llama4_2025}, Qwen VL(3~\&~3.5)~\citep{bai2025qwen3vltechnicalreport}, Gemma 3~\citep{gemmateam2025gemma3technicalreport}, InternVL 3.5~\citep{wang2025internvl35advancingopensourcemultimodal}, SEA-LION VL~\citep{aisingapore2026sealionv4}, and Molmo 2~\citep{clark2026molmo2openweightsdata}. We employed prompt engineering to instruct the model to generate responses in JSON format, which consists of step-by-step solution and a concise final answer. We subsequently parsed the JSON outputs to extract both the reasoning trace and the final answer. In cases when a model failed to generate the requested format, we implemented regex-based pattern matching as a fallback parsing mechanism. To account for minor deviations in numerical short answers, we accept values within a 1\% tolerance of the ground truth. More details regarding the experiment setup are available in Appendix~\ref{appendix:c}.

To evaluate model performance, we adopt metrics introduced by \cite{zou2025dynamathdynamicvisualbenchmark}: \textbf{average-case accuracy} ($\mathcal{A}_{avg}$) and \textbf{worst-case accuracy} ($\mathcal{A}_{wst}$), which defined as follows:

\begin{equation}
\begin{aligned}
\label{eq:avg_wst_define}
    \mathcal{A}_{avg} = \frac{1}{N} \sum_{i=1}^{N} \frac{1}{M} \sum_{j=1}^{M} \mathbb{I}[\text{Ans}(i,j) = \text{GT}(i,j)],  \ 
    \mathcal{A}_{wst} = \frac{1}{N} \sum_{i=1}^{N} \min_{j \in [1,M]} \mathbb{I}[\text{Ans}(i,j) = \text{GT}(i,j)], 
\end{aligned}
\end{equation}
where $\text{Ans}(i,j)$ and $\text{GT}(i,j)$ represent the generated answer and the ground truth answer, respectively, for variant $j$ of question $i$.


\begin{table}[ht]
\centering
\caption{English-only results on \datasetname{}. For each subject, $A_{avg}$ denotes average-case accuracy and $A_{wst}$ denotes worst-case accuracy. $\Delta$ denotes the reduction $A_{wst} - A_{avg}$.}
\label{tab:stem_vision_english_avg_wc_by_subject}
\resizebox{\textwidth}{!}{%
\begin{tabular}{l|rrc|rrc|rrc|rrc|rrc}
\hline
\textbf{Model} & \multicolumn{3}{c}{\textbf{Mathematics}} & \multicolumn{3}{c}{\textbf{Physics}} & \multicolumn{3}{c}{\textbf{Chemistry}} & \multicolumn{3}{c}{\textbf{Biology}} & \multicolumn{3}{c}{\textbf{Compsci}} \\
 & $A_{avg}$ & $A_{wst}$ & $\Delta$ & $A_{avg}$ & $A_{wst}$ & $\Delta$ & $A_{avg}$ & $A_{wst}$ & $\Delta$ & $A_{avg}$ & $A_{wst}$ & $\Delta$ & $A_{avg}$ & $A_{wst}$ & $\Delta$ \\
\hline
\multicolumn{16}{c}{\textit{Closed-source Models}} \\
\hline
GPT-5.4 & 90.3 & 72.4 & -17.9 & 94.6 & 84.6 & -10.0 & 94.4 & 78.8 & -15.6 & 72.2 & 47.8 & -24.4 & 95.4 & 84.6 & -10.8 \\
Gemini-2.5-Pro & 86.2 & 66.3 & -19.9 & 90.4 & 69.2 & -21.2 & 91.7 & 70.1 & -21.6 & 71.7 & 43.5 & -28.2 & 91.2 & 75.0 & -16.2 \\
Gemini-2.5-Flash & 85.2 & 61.8 & -23.4 & 87.4 & 63.1 & -24.3 & 92.2 & 67.9 & -24.3 & 51.7 & 30.4 & -21.3 & 86.9 & 63.5 & -23.4 \\
\hline
\multicolumn{16}{l}{} \\
\multicolumn{16}{c}{\textit{Open-source Models}} \\
\hline
Qwen3.5-122B-A10B & 82.1 & 49.2 & -32.9 & 86.7 & 49.2 & -37.5 & 88.0 & 43.1 & -44.9 & 67.8 & 26.1 & -41.7 & 88.5 & 57.7 & -30.8 \\
Qwen3.5-27B & 85.1 & 60.3 & -24.8 & 87.0 & 56.4 & -30.6 & 90.9 & 65.7 & -25.2 & 65.7 & 39.1 & -26.6 & 89.2 & 63.5 & -25.7 \\
Llama-4-Scout & 71.4 & 33.7 & -37.7 & 78.8 & 43.1 & -35.7 & 84.7 & 46.7 & -38.0 & 44.3 & 8.7 & -35.6 & 68.7 & 25.0 & -43.7 \\
InternVL3.5-14B & 58.5 & 21.1 & -37.4 & 69.2 & 30.3 & -38.9 & 76.8 & 35.0 & -41.8 & 30.9 & 4.3 & -26.6 & 64.6 & 21.2 & -43.4 \\
InternVL3.5-8B & 64.8 & 35.2 & -29.6 & 73.8 & 45.1 & -28.7 & 79.3 & 40.9 & -38.4 & 36.5 & 4.3 & -32.2 & 61.0 & 25.0 & -36.0 \\
InternVL3.5-4B & 54.0 & 21.1 & -32.9 & 65.3 & 26.7 & -38.6 & 72.6 & 29.9 & -42.7 & 23.9 & 0.0 & -23.9 & 55.2 & 17.3 & -37.9 \\
Qwen3-VL-8B-Instruct & 61.5 & 25.6 & -35.9 & 70.3 & 34.4 & -35.9 & 83.6 & 47.4 & -36.2 & 53.5 & 8.7 & -44.8 & 67.1 & 25.0 & -42.1 \\
Qwen3-VL-4B-Instruct & 31.7 & 5.0 & -26.7 & 40.4 & 6.2 & -34.2 & 35.8 & 3.6 & -32.2 & 17.8 & 0.0 & -17.8 & 23.8 & 3.8 & -20.0 \\
Molmo2-8B & 46.5 & 14.6 & -31.9 & 61.9 & 30.3 & -31.6 & 56.6 & 24.1 & -32.5 & 26.1 & 8.7 & -17.4 & 40.8 & 9.6 & -31.2 \\
Molmo2-4B & 40.9 & 15.6 & -25.3 & 54.6 & 25.6 & -29.0 & 49.6 & 22.6 & -27.0 & 16.5 & 0.0 & -16.5 & 31.5 & 7.7 & -23.8 \\
Gemma-3-12B-IT & 61.7 & 29.1 & -32.6 & 65.4 & 33.3 & -32.1 & 60.1 & 26.3 & -33.8 & 34.3 & 8.7 & -25.6 & 65.6 & 17.3 & -48.3 \\
Gemma-3-4B-IT & 46.2 & 14.1 & -32.1 & 44.3 & 12.8 & -31.5 & 43.4 & 8.8 & -34.6 & 16.1 & 0.0 & -16.1 & 36.5 & 5.8 & -30.7 \\
SEA-LION-v4-8B-VL & 68.2 & 32.2 & -36.0 & 75.6 & 44.6 & -31.0 & 83.4 & 49.6 & -33.8 & 51.7 & 4.3 & -47.4 & 71.7 & 36.5 & -35.2 \\
SEA-LION-v4-4B-VL & 53.0 & 12.1 & -40.9 & 53.6 & 16.9 & -36.7 & 63.6 & 17.5 & -46.1 & 26.5 & 0.0 & -26.5 & 44.2 & 11.5 & -32.7 \\
\hline
\end{tabular}%
}
\end{table}

\begin{table}[ht]
\centering
\caption{Reasoning step performance. Subjects are abbreviated as follows: Math = Mathematics, Phys = Physics, Chem = Chemistry, Bio = Biology, CS = Computer Science. The Pearson correlation between $\mathcal{P}$ and $\mathcal{R}$ is $r = 0.959$.}
\label{tab:reasoning_judge_subject_metrics}
\resizebox{\textwidth}{!}{%
\begin{tabular}{l|rrrr|rrrr|rrrr|rrrr|rrrr}
\hline
\textbf{Subject} & \multicolumn{4}{c}{\textbf{GPT-5.4}} & \multicolumn{4}{c}{\textbf{InternVL3.5-8B}} & \multicolumn{4}{c}{\textbf{Qwen3-VL-8B}} & \multicolumn{4}{c}{\textbf{Molmo2-8B}} & \multicolumn{4}{c}{\textbf{Gemma-3-4B-IT}} \\
 & \textbf{$\mathcal{P}$} & \textbf{$\mathcal{R}$} & \textbf{$\mathcal{F}$} & \textbf{$\mathcal{F}_{wst}$} & \textbf{$\mathcal{P}$} & \textbf{$\mathcal{R}$} & \textbf{$\mathcal{F}$} & \textbf{$\mathcal{F}_{wst}$} & \textbf{$\mathcal{P}$} & \textbf{$\mathcal{R}$} & \textbf{$\mathcal{F}$} & \textbf{$\mathcal{F}_{wst}$} & \textbf{$\mathcal{P}$} & \textbf{$\mathcal{R}$} & \textbf{$\mathcal{F}$} & \textbf{$\mathcal{F}_{wst}$} & \textbf{$\mathcal{P}$} & \textbf{$\mathcal{R}$} & \textbf{$\mathcal{F}$} & \textbf{$\mathcal{F}_{wst}$} \\
\hline
Math & 91.1 & 87.8 & 88.0 & 67.8 & 67.7 & 69.8 & 66.2 & 39.9 & 64.2 & 65.1 & 62.7 & 30.1 & 57.6 & 53.6 & 51.5 & 23.6 & 52.3 & 58.9 & 51.2 & 20.1 \\
Phys & 95.0 & 91.7 & 92.2 & 79.2 & 78.7 & 79.5 & 76.6 & 53.9 & 77.6 & 78.2 & 75.9 & 51.7 & 72.8 & 73.0 & 70.2 & 44.0 & 53.7 & 61.0 & 51.4 & 21.4 \\
Chem & 94.3 & 94.1 & 93.4 & 79.0 & 81.6 & 84.1 & 80.7 & 52.5 & 83.7 & 86.1 & 83.5 & 53.4 & 76.3 & 73.5 & 71.9 & 39.4 & 59.3 & 65.0 & 57.6 & 20.7 \\
Bio & 81.8 & 82.1 & 81.1 & 60.3 & 51.0 & 54.0 & 49.5 & 24.0 & 60.3 & 61.7 & 57.9 & 22.9 & 51.4 & 47.9 & 47.0 & 20.6 & 30.1 & 36.3 & 28.6 & 7.7 \\
CS & 94.3 & 88.2 & 89.4 & 75.1 & 68.7 & 61.4 & 61.3 & 32.4 & 67.9 & 64.4 & 63.4 & 33.2 & 53.9 & 42.0 & 42.6 & 15.5 & 48.9 & 41.2 & 39.6 & 14.1 \\
\hline
\end{tabular}%
}
\end{table}

Additionally, to evaluate step-level reasoning quality, we decompose performance into \textbf{Recall} ($\mathcal{R}$), which measures whether a model covers all necessary reasoning steps, and \textbf{Precision} ($\mathcal{P}$), which measures whether model-generated steps are valid, i.e., correspond to correct reasoning rather than redundant, hallucinated, or incorrect ones, inspired by \cite{xie2026finchainsymbolicbenchmarkverifiable} and  \cite{gull2026engtracesymbolicbenchmarkverifiable}. To automate this process, we employ GPT-5.4 as an LLM judge $\mathcal{J}$ (the full prompt template is provided in Appendix~\ref{appendix:c2}) to provide binary step-level decisions by comparing the model's reasoning against a ground-truth reference. For each problem instance $i$, let the reference solution be 
$\mathcal{S}_{\text{gold}}^{(i)} = \{g_1^{(i)}, \dots, g_m^{(i)}\}$ 
and the model output be 
$\mathcal{S}_{\text{model}}^{(i)} = \{s_1^{(i)}, \dots, s_n^{(i)}\}$. The judge's evaluations are defined as follows:

\begin{equation}
\label{eq:judge_define}
\mathcal{J}_{\text{rec}}(g, \mathcal{S}_{\text{model}}^{(i)}) = 
\begin{cases} 
1 & \text{if } g \text{ is covered by } \mathcal{S}_{\text{model}}^{(i)} \\
0 & \text{otherwise}
\end{cases}, \qquad
\mathcal{J}_{\text{pre}}(s, \mathcal{S}_{\text{gold}}^{(i)}) = 
\begin{cases} 
1 & \text{if } s \text{ is valid w.r.t } \mathcal{S}_{\text{gold}}^{(i)} \\
0 & \text{otherwise}
\end{cases}
\end{equation}

Instance-level metrics for a problem $i$ are then calculated as:
\begin{equation}
\label{eq:metrics_define}
\mathcal{R}_i = \frac{1}{m} \sum_{k=1}^{m} 
\mathcal{J}_{\text{rec}}(g_k^{(i)}, \mathcal{S}_{\text{model}}^{(i)}), \quad
\mathcal{P}_i = \frac{1}{n} \sum_{l=1}^{n} 
\mathcal{J}_{\text{pre}}(s_l^{(i)}, \mathcal{S}_{\text{gold}}^{(i)}), \quad
\mathcal{F}_i = \frac{2 \cdot \mathcal{P}_i \cdot \mathcal{R}_i}{\mathcal{P}_i + \mathcal{R}_i + \epsilon}
\end{equation}
To evaluate consistency across $M$ variants of a problem, we report the \textbf{Worst-case Step F1}:
\begin{equation}
\label{eq:f1_wst_define}
    \mathcal{F}_{wst} = \frac{1}{N} \sum_{i=1}^{N} \min_{j \in [1,M]} \mathcal{F}_{i,j}
\end{equation}


\subsection{Main Result} \label{sec:main_result}

\paragraph{English-only Result}

Table~\ref{tab:stem_vision_english_avg_wc_by_subject} presents the experimental results for the English subset of \datasetname{}. Among all evaluated models, \textsc{GPT-5.4} achieves the highest performance, securing over 90\% average accuracy in Mathematics, Physics, Chemistry, and Computer Science, with a minimum of 72.2\% in Biology. Among the open-weight models, Qwen3.5-27B yields the best results at 65-90\%, almost comparable to Gemini-2.5-Pro. Interestingly, Qwen3.5-27B outperforms Qwen3.5-122B-A10B in almost all subjects, likely due to the latter's Mixture-of-Experts (MoE) architecture, which only activates 10B parameters during inference. In almost every subject, SEA-LION-v4-8B-VL is the strongest model among 8-14B variants, securing an average accuracy of 51.7\% in Biology and 68-83\% in other subjects. Among 4B models, InternVL3.5-4B ranks highest, with 23.9\% average accuracy in Biology and 54-72\% in others. These results highlight the significant difficulty of the Biology subset compared to other subjects. With the exception of InternVL3.5-14B, model performance within the same family consistently scales with size.

Despite remarkable average performance across the dataset, frontier models struggle to solve problem variants consistently, exhibiting performance drops ($\Delta$) ranging from -10\% to -28\% in closed-weight models. Smaller open-weight models suffer even more severe degradation. For instance, Qwen3.5-122B drops sharply in Chemistry ($\Delta = -44.9$) and Biology ($\Delta = -41.7$). SEA-LION-v4-8B-VL nearly collapses in Biology (51.7\% $\to$ 4.3\%, $\Delta = -47.4$), while Gemma-3-12B-IT suffers a steep decline in Computer Science (65.6\% $\to$ 17.3\%, $\Delta = -48.3$). Notably, under worst-case conditions, several smaller models such as InternVL3.5-4B and Molmo2-4B break down entirely in Biology, with $\Delta = -23.9$ and $\Delta = -16.5$, respectively. Furthermore, despite achieving similar average accuracy, Qwen3.5-122B-A10B has a much larger $\Delta$ across all subjects compared to Qwen3.5-27B, resulting in a substantially lower worst-case accuracy. These findings expose the inconsistency of current VLMs across problem variants, underscoring the critical need to train more robust models capable of reliable reasoning in diverse scenarios.

\begin{figure}
    \centering
    \includegraphics[width=1.0\linewidth]{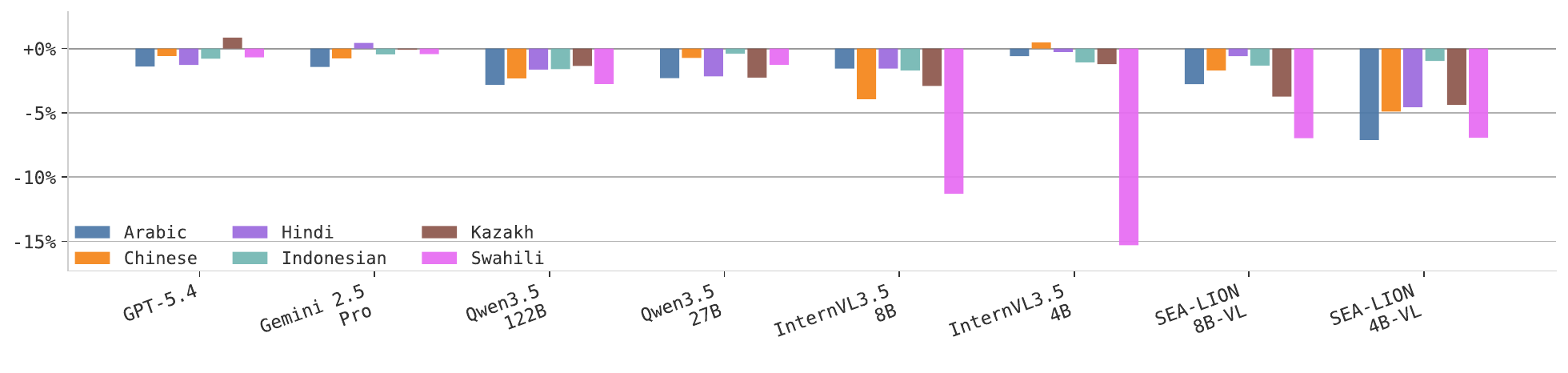}
    \caption{Average-case accuracy gap ($\Delta \mathcal{A}_{avg}$) relative to English across languages and models. Negative values indicate performance degradation in non-English settings.}
    \label{fig:multilingual_compare}
    \vspace{-0.5cm}
\end{figure}

\paragraph{Multilingual Results}
Figure~\ref{fig:multilingual_compare} compares VLM performance across seven languages by reporting $\Delta \mathcal{A}_{avg}$ relative to English. Overall, language variants have minimal effect on proprietary and large open-weight models, with GPT-5.4 and Gemini 2.5 showing near-zero degradation across all tested languages. In contrast, smaller open-weight models exhibit notable performance drops in non-English settings, with degradation becoming more pronounced in lower-resource languages. For example, Swahili exhibits the most substantial accuracy gap across smaller models, reflecting its underrepresentation in pretraining corpora. By comparison, higher-resource languages such as Chinese and Indonesian demonstrate greater resilience, showing only mild degradation.

\paragraph{Reasoning Step Performance}
To analyze reasoning quality of the models, we conduct step-level evaluation. Due to constrained budget, we selected a representative subset of 
five models covering a broad range of performance levels reported in 
Table~\ref{tab:stem_vision_english_avg_wc_by_subject}. Observed from Table~\ref{tab:reasoning_judge_subject_metrics}, across all evaluated models and subjects, $\mathcal{P}$ and $\mathcal{R}$ exhibit a very high correlation, suggesting that reasoning quality operates as a general capability rather than a precision–recall tradeoff, with both metrics largely reflecting the model’s underlying reasoning correctness rather than separate tendencies to over- or under-generate steps.

A consistent gap between $\mathcal{F}$ and $\mathcal{F}_\text{wst}$ across all 
evaluated models reveals reasoning step fragility, most pronounced in Mathematics and 
Computer Science. Notably, even GPT-5.4 drops 
from $\mathcal{F}$ to $\mathcal{F}_\text{wst}$ in the range of $13.0$ to 
$20.2$ points across subjects, suggesting that worst-case reasoning fragility is a 
general phenomenon rather than a weakness of smaller models alone. This degradation 
worsens for weaker models, Gemma-3-4B-IT degrades from $\mathcal{F} {=} 39.6$ to 
$\mathcal{F}_\text{wst} {=} 14.1$ in Computer Science and from $\mathcal{F} {=} 28.6$ 
to $\mathcal{F}_\text{wst} {=} 7.7$ in Biology, suggesting that average F1 alone 
may overestimate true reasoning capability. At the subject level, Mathematics 
exhibits the largest $\mathcal{F}$-to-$\mathcal{F}_\text{wst}$ drop across nearly 
all models, while Chemistry and Physics consistently yield stronger step-level 
performance. Biology remains the most challenging across all 
evaluated models, aligning with the accuracy trends in 
Table~\ref{tab:stem_vision_english_avg_wc_by_subject}.


\begin{figure}
    \centering
    \includegraphics[width=1.0\linewidth]{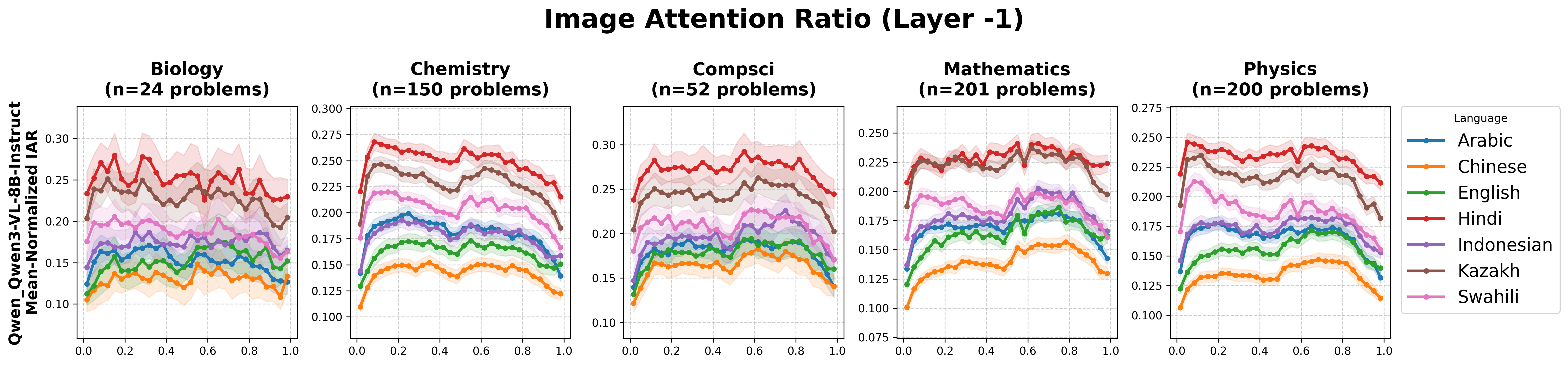}
    \caption{IAR trends across subjects and languages for \textit{Qwen3-VL-8B-Instruct} (95\% confidence intervals are shown).}
    \vspace{-0.5cm}
    \label{fig:iar}
\end{figure}

\subsection{Attention Heads Analysis} \label{sec:attention_heads_analysis}
To gain deeper insights into the model's internal reasoning during response generation, we examined the \textit{Qwen3-VL-8B-Instruct}'s attention heads at the final layer. Inspired by previous study on VLMs modality bias~\citep{chen2025truemultimodalincontextlearning, deng2025wordsvisionvisionlanguagemodels, zheng2025mllmsdeeplyaffectedmodality} and attention heads inspection during output generation~\citep{yin2025refusalfallscliffsafety}, this experiment quantifies the extent to which the model attends to visual tokens relative to textual tokens.

When generating token $t$, we first average the attention weights of all previous context tokens across all $H$ heads. Then, we calculate the average attention directed towards the image tokens ($I$) and the average attention directed towards the text tokens ($T$):

$$ A_{I} = \frac{1}{|I|} \sum_{i \in I} \bar{a}_{t, i} \quad \text{and} \quad A_{T} = \frac{1}{|T|} \sum_{k \in T} \bar{a}_{t, k} $$

where $\bar{a}_{t, i}$ and $\bar{a}_{t, k}$ represent the head-averaged attention weights from the generated token $t$ to image token $i$ and text token $k$, respectively, and $|I|$ and $|T|$ denote the total number of tokens within the image and text sets. Finally, we calculate the \textbf{Image Attention Ratio} (IAR) by dividing the average image attention by the sum of both:

$$ \text{IAR} = \frac{A_{I}}{A_{I} + A_{T}} $$

In our experiment, we excluded special tokens (e.g., the start token), the prompt template, and newly generated response tokens to ensure only tokens related to the problem were measured. We calculated the IAR for the entire response to measure the trend of image-component utilization as text generation proceeds. We then normalized the length of all responses to a scale of $0$ to $1$. Finally, we averaged the results across all problems within the same subject.

Figure~\ref{fig:iar} illustrates the IAR trends across the five subjects. The results demonstrate that the IAR varies across languages, with Chinese exhibiting the lowest ratio. This indicates that the model relies less on the visual context when prompted in Chinese. In contrast, queries in Kazakh and Hindi yield the highest IARs, indicating a much stronger reliance on the image. These findings highlight a difference in how the model balances text and image information across languages. We hypothesize that higher language familiarity allows the model to rely more on text part of the inputs, whereas lower familiarity forces it to compensate by depending more heavily on visual inputs.

\subsection{Additional Analysis} \label{sec:qualitative_analysis}
\paragraph{Exact Computation vs Visual Intuition}
We observed a notable failure case where the frontier model, GPT-5.4, preferred to rely on visual intuition rather than computing the exact answer, as illustrated in Figure~\ref{fig:sample_fail}. In problem $Mat020$ (variant $04$), the model successfully computed the answer based on the provided numerical information. However, in variant $07$, the model initially follows the correct reasoning path to derive the exact answer, but is then distracted by the accompanying figure, which visually depicts point $T$ located close to $x=0$. This ultimately leads to an incorrect final answer. This behavior highlights that, at boundary conditions, visual cues can override a model's logical reasoning and distract it from computing the exact solution.

\begin{figure}[t]
    \centering
    \begin{subfigure}[b]{0.42\textwidth}
        \centering
        \includegraphics[width=0.8\linewidth]{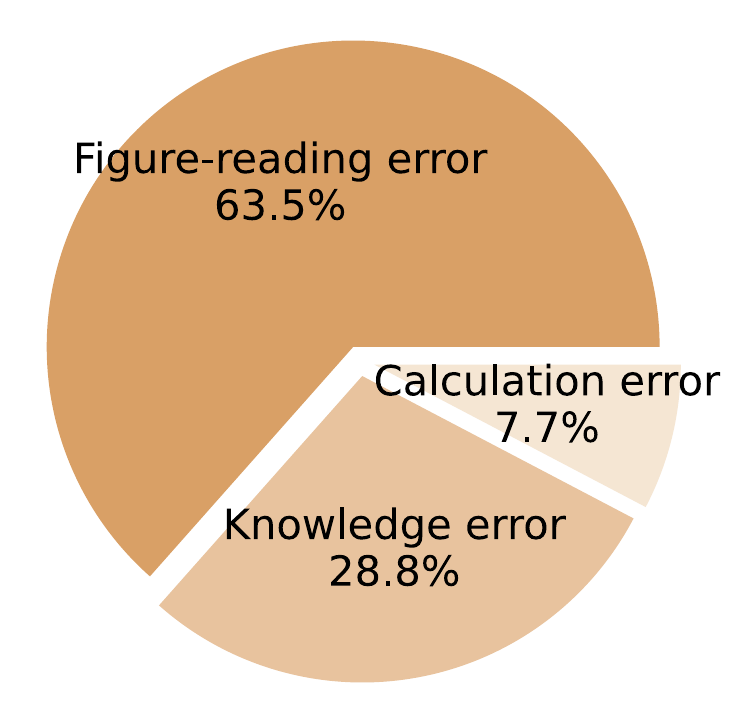}
        \caption{Distribution of error types.}
        \label{fig:error_analysis_pie}
    \end{subfigure}
    \hfill
    \begin{subfigure}[b]{0.55\textwidth}
        \centering
        \includegraphics[width=\linewidth]{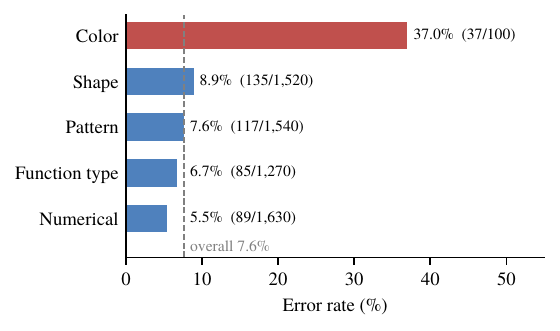}
        \caption{Error rate by perturbation type.}
        \label{fig:perturbation_breakdown}
    \end{subfigure}
    \caption{Error analysis of GPT-5.4 on the English subset. (a) Distribution of error types across 52 manually inspected templates. (b) Error rate for each perturbation type, with incorrect answers over total instances in parentheses. The dashed line marks the overall error rate.}
    \label{fig:error_analysis}
    \vspace{-3mm}
\end{figure}

\paragraph{Failure Mode Analysis}
To better understand how VLMs fail to answer our problems, we conducted a manual inspection of 52 templates where GPT-5.4 produced incorrect answers. We categorized the failures into three common error types: (1) \textbf{figure-reading errors} occur when the VLM fails to extract accurate numerical values or miscounts number of objects within the image (2) \textbf{knowledge errors} happen when the model demonstrates flawed logical reasoning, such as using inapplicable theorems, declaring unstated assumptions, or hallucinating non-existent information (3) \textbf{calculation errors} occur when the VLM fails to perform numerical computations correctly, despite following the correct underlying logic and steps. Figure~\ref{fig:error_analysis_pie} illustrates the distribution of these errors. Figure-reading errors are the most frequent, accounting for 63.5\%, followed by knowledge errors at 28.8\%, with calculation errors comprising the remaining 7.7\%. These findings emphasize that future advancements in VLM development must prioritize robust architectural designs and training methodologies that enhance precise visual grounding and mitigate hallucinations.


\paragraph{Error Breakdown by Perturbation Type}
To understand which variations drive the gap between average-case and worst-case accuracy, we computed the error rate of GPT-5.4 for each perturbation type on the English subset (Figure~\ref{fig:perturbation_breakdown}). Color perturbations yield the highest error rate at 37.0\%, although only 100 instances feature Color variations. All other types stay below 10\%, with Numerical variations lowest at 5.5\%. Shape and Pattern variations together account for 252 of the 463 incorrect answers (54.4\%), in line with figure-reading errors being the most frequent error type (Figure~\ref{fig:error_analysis_pie}).



\section{Conclusion} \label{sec:conclusion}
Our work introduce \datasetname{} as a visually-grounded symbolic benchmark in five STEM subjects and seven languages. We collect responses from 17 VLMs and evaluate their performance using six metrics. Our experiments reveal a noticeable performance gap between average-case and worst-case accuracy across all models and subjects, while the effect of multilinguality is only apparent in small open-weight models. Similar performance trends are also observed when the models are evaluated based on their reasoning steps. Finally, we discover cross-lingual variation in the relative attention allocated to image tokens compared to text tokens when inspecting the attention heads of \textit{Qwen3-VL-8B-Instruct}. Our work highlights the importance of evaluation beyond static benchmarks for measuring the quality of VLMs.

\subsection*{AI use statement}


In this work, we used generative AI tools for generate synthetic datasets, design or provide feedback on research methodology or experiments, implement methods, assist with translation, clean and reformat dataset, support qualitative and thematic data analysis. We have not used generative AI tools for interpret results, and “help develop theoretical models or conceptual frameworks, formulate mathematical claims, provide critical ingredients for proving mathematical claims, assist in the writing of proofs, propose or refine hypotheses” are not applicable to this work. Additionally, we used generative AI tools for create or modify scientific figures or images, suggest experimental parameters, create or edit software code, creation of artifacts, draft parts of a research paper, summarize or analyze existing literature, discover research topics or identify gaps, brainstorming, sourcing/searching for information, edit a research paper to improve readability, and identify relevant literature. We have reviewed all AI-assisted work. The AI-translated dataset is verified by native speakers of corresponding languages. We only used AI to find prior works and manually draft the related works section to prevent the possibility of AI hallucination. We take responsibility for the final content of this work, including text, claims or artifacts produced with the aid of generative AI.

\bibliography{iclr2027_conference}

@misc{wang2024measuringmultimodalmathematicalreasoning,
      title={Measuring Multimodal Mathematical Reasoning with MATH-Vision Dataset}, 
      author={Ke Wang and Junting Pan and Weikang Shi and Zimu Lu and Mingjie Zhan and Hongsheng Li},
      year={2024},
      eprint={2402.14804},
      archivePrefix={arXiv},
      primaryClass={cs.CV},
      url={https://arxiv.org/abs/2402.14804}, 
}

@InProceedings{Irawan_2026_CVPR,
    author    = {Irawan, Patrick Amadeus and Hanif, Ikhlasul Akmal and Al Kautsar, Muhammad Dehan and Winata, Genta Indra and Koto, Fajri and Aji, Alham Fikri},
    title     = {Vision Language Models are Confused Tourists},
    booktitle = {Proceedings of the IEEE/CVF Conference on Computer Vision and Pattern Recognition (CVPR) Findings},
    month     = {June},
    year      = {2026},
    pages     = {1763-1773}
}

@misc{vo2026visionlanguagemodelsbiased,
      title={Vision Language Models are Biased}, 
      author={An Vo and Khai-Nguyen Nguyen and Mohammad Reza Taesiri and Vy Tuong Dang and Anh Totti Nguyen and Daeyoung Kim},
      year={2026},
      eprint={2505.23941},
      archivePrefix={arXiv},
      primaryClass={cs.LG},
      url={https://arxiv.org/abs/2505.23941}, 
}

@misc{xie2026finchainsymbolicbenchmarkverifiable,
      title={FinChain: A Symbolic Benchmark for Verifiable Chain-of-Thought Financial Reasoning}, 
      author={Zhuohan Xie and Daniil Orel and Rushil Thareja and Dhruv Sahnan and Hachem Madmoun and Fan Zhang and Debopriyo Banerjee and Georgi Georgiev and Xueqing Peng and Lingfei Qian and Jimin Huang and Jinyan Su and Aaryamonvikram Singh and Rui Xing and Rania Elbadry and Chen Xu and Haonan Li and Fajri Koto and Ivan Koychev and Tanmoy Chakraborty and Yuxia Wang and Salem Lahlou and Veselin Stoyanov and Sophia Ananiadou and Preslav Nakov},
      year={2026},
      eprint={2506.02515},
      archivePrefix={arXiv},
      primaryClass={cs.CL},
      url={https://arxiv.org/abs/2506.02515}, 
}

@misc{meta_llama4_2025,
  author       = {{Meta AI}},
  title        = {Introducing LLaMA 4},
  year         = {2025},
  url          = {https://ai.meta.com/llama/},
  note         = {Accessed: 2026-04-16}
}

@misc{singh2025openaigpt5card,
      title={OpenAI GPT-5 System Card}, 
      author={Aaditya Singh and Adam Fry and Adam Perelman and Adam Tart and Adi Ganesh and Ahmed El-Kishky and Aidan McLaughlin and Aiden Low and AJ Ostrow and Akhila Ananthram and Akshay Nathan and Alan Luo and Alec Helyar and Aleksander Madry and Aleksandr Efremov and Aleksandra Spyra and Alex Baker-Whitcomb and Alex Beutel and Alex Karpenko and Alex Makelov and Alex Neitz and Alex Wei and Alexandra Barr and Alexandre Kirchmeyer and Alexey Ivanov and Alexi Christakis and Alistair Gillespie and Allison Tam and Ally Bennett and Alvin Wan and Alyssa Huang and Amy McDonald Sandjideh and Amy Yang and Ananya Kumar and Andre Saraiva and Andrea Vallone and Andrei Gheorghe and Andres Garcia Garcia and Andrew Braunstein and Andrew Liu and Andrew Schmidt and Andrey Mereskin and Andrey Mishchenko and Andy Applebaum and Andy Rogerson and Ann Rajan and Annie Wei and Anoop Kotha and Anubha Srivastava and Anushree Agrawal and Arun Vijayvergiya and Ashley Tyra and Ashvin Nair and Avi Nayak and Ben Eggers and Bessie Ji and Beth Hoover and Bill Chen and Blair Chen and Boaz Barak and Borys Minaiev and Botao Hao and Bowen Baker and Brad Lightcap and Brandon McKinzie and Brandon Wang and Brendan Quinn and Brian Fioca and Brian Hsu and Brian Yang and Brian Yu and Brian Zhang and Brittany Brenner and Callie Riggins Zetino and Cameron Raymond and Camillo Lugaresi and Carolina Paz and Cary Hudson and Cedric Whitney and Chak Li and Charles Chen and Charlotte Cole and Chelsea Voss and Chen Ding and Chen Shen and Chengdu Huang and Chris Colby and Chris Hallacy and Chris Koch and Chris Lu and Christina Kaplan and Christina Kim and CJ Minott-Henriques and Cliff Frey and Cody Yu and Coley Czarnecki and Colin Reid and Colin Wei and Cory Decareaux and Cristina Scheau and Cyril Zhang and Cyrus Forbes and Da Tang and Dakota Goldberg and Dan Roberts and Dana Palmie and Daniel Kappler and Daniel Levine and Daniel Wright and Dave Leo and David Lin and David Robinson and Declan Grabb and Derek Chen and Derek Lim and Derek Salama and Dibya Bhattacharjee and Dimitris Tsipras and Dinghua Li and Dingli Yu and DJ Strouse and Drew Williams and Dylan Hunn and Ed Bayes and Edwin Arbus and Ekin Akyurek and Elaine Ya Le and Elana Widmann and Eli Yani and Elizabeth Proehl and Enis Sert and Enoch Cheung and Eri Schwartz and Eric Han and Eric Jiang and Eric Mitchell and Eric Sigler and Eric Wallace and Erik Ritter and Erin Kavanaugh and Evan Mays and Evgenii Nikishin and Fangyuan Li and Felipe Petroski Such and Filipe de Avila Belbute Peres and Filippo Raso and Florent Bekerman and Foivos Tsimpourlas and Fotis Chantzis and Francis Song and Francis Zhang and Gaby Raila and Garrett McGrath and Gary Briggs and Gary Yang and Giambattista Parascandolo and Gildas Chabot and Grace Kim and Grace Zhao and Gregory Valiant and Guillaume Leclerc and Hadi Salman and Hanson Wang and Hao Sheng and Haoming Jiang and Haoyu Wang and Haozhun Jin and Harshit Sikchi and Heather Schmidt and Henry Aspegren and Honglin Chen and Huida Qiu and Hunter Lightman and Ian Covert and Ian Kivlichan and Ian Silber and Ian Sohl and Ibrahim Hammoud and Ignasi Clavera and Ikai Lan and Ilge Akkaya and Ilya Kostrikov and Irina Kofman and Isak Etinger and Ishaan Singal and Jackie Hehir and Jacob Huh and Jacqueline Pan and Jake Wilczynski and Jakub Pachocki and James Lee and James Quinn and Jamie Kiros and Janvi Kalra and Jasmyn Samaroo and Jason Wang and Jason Wolfe and Jay Chen and Jay Wang and Jean Harb and Jeffrey Han and Jeffrey Wang and Jennifer Zhao and Jeremy Chen and Jerene Yang and Jerry Tworek and Jesse Chand and Jessica Landon and Jessica Liang and Ji Lin and Jiancheng Liu and Jianfeng Wang and Jie Tang and Jihan Yin and Joanne Jang and Joel Morris and Joey Flynn and Johannes Ferstad and Johannes Heidecke and John Fishbein and John Hallman and Jonah Grant and Jonathan Chien and Jonathan Gordon and Jongsoo Park and Jordan Liss and Jos Kraaijeveld and Joseph Guay and Joseph Mo and Josh Lawson and Josh McGrath and Joshua Vendrow and Joy Jiao and Julian Lee and Julie Steele and Julie Wang and Junhua Mao and Kai Chen and Kai Hayashi and Kai Xiao and Kamyar Salahi and Kan Wu and Karan Sekhri and Karan Sharma and Karan Singhal and Karen Li and Kenny Nguyen and Keren Gu-Lemberg and Kevin King and Kevin Liu and Kevin Stone and Kevin Yu and Kristen Ying and Kristian Georgiev and Kristie Lim and Kushal Tirumala and Kyle Miller and Lama Ahmad and Larry Lv and Laura Clare and Laurance Fauconnet and Lauren Itow and Lauren Yang and Laurentia Romaniuk and Leah Anise and Lee Byron and Leher Pathak and Leon Maksin and Leyan Lo and Leyton Ho and Li Jing and Liang Wu and Liang Xiong and Lien Mamitsuka and Lin Yang and Lindsay McCallum and Lindsey Held and Liz Bourgeois and Logan Engstrom and Lorenz Kuhn and Louis Feuvrier and Lu Zhang and Lucas Switzer and Lukas Kondraciuk and Lukasz Kaiser and Manas Joglekar and Mandeep Singh and Mandip Shah and Manuka Stratta and Marcus Williams and Mark Chen and Mark Sun and Marselus Cayton and Martin Li and Marvin Zhang and Marwan Aljubeh and Matt Nichols and Matthew Haines and Max Schwarzer and Mayank Gupta and Meghan Shah and Melody Huang and Meng Dong and Mengqing Wang and Mia Glaese and Micah Carroll and Michael Lampe and Michael Malek and Michael Sharman and Michael Zhang and Michele Wang and Michelle Pokrass and Mihai Florian and Mikhail Pavlov and Miles Wang and Ming Chen and Mingxuan Wang and Minnia Feng and Mo Bavarian and Molly Lin and Moose Abdool and Mostafa Rohaninejad and Nacho Soto and Natalie Staudacher and Natan LaFontaine and Nathan Marwell and Nelson Liu and Nick Preston and Nick Turley and Nicklas Ansman and Nicole Blades and Nikil Pancha and Nikita Mikhaylin and Niko Felix and Nikunj Handa and Nishant Rai and Nitish Keskar and Noam Brown and Ofir Nachum and Oleg Boiko and Oleg Murk and Olivia Watkins and Oona Gleeson and Pamela Mishkin and Patryk Lesiewicz and Paul Baltescu and Pavel Belov and Peter Zhokhov and Philip Pronin and Phillip Guo and Phoebe Thacker and Qi Liu and Qiming Yuan and Qinghua Liu and Rachel Dias and Rachel Puckett and Rahul Arora and Ravi Teja Mullapudi and Raz Gaon and Reah Miyara and Rennie Song and Rishabh Aggarwal and RJ Marsan and Robel Yemiru and Robert Xiong and Rohan Kshirsagar and Rohan Nuttall and Roman Tsiupa and Ronen Eldan and Rose Wang and Roshan James and Roy Ziv and Rui Shu and Ruslan Nigmatullin and Saachi Jain and Saam Talaie and Sam Altman and Sam Arnesen and Sam Toizer and Sam Toyer and Samuel Miserendino and Sandhini Agarwal and Sarah Yoo and Savannah Heon and Scott Ethersmith and Sean Grove and Sean Taylor and Sebastien Bubeck and Sever Banesiu and Shaokyi Amdo and Shengjia Zhao and Sherwin Wu and Shibani Santurkar and Shiyu Zhao and Shraman Ray Chaudhuri and Shreyas Krishnaswamy and Shuaiqi and Xia and Shuyang Cheng and Shyamal Anadkat and Simón Posada Fishman and Simon Tobin and Siyuan Fu and Somay Jain and Song Mei and Sonya Egoian and Spencer Kim and Spug Golden and SQ Mah and Steph Lin and Stephen Imm and Steve Sharpe and Steve Yadlowsky and Sulman Choudhry and Sungwon Eum and Suvansh Sanjeev and Tabarak Khan and Tal Stramer and Tao Wang and Tao Xin and Tarun Gogineni and Taya Christianson and Ted Sanders and Tejal Patwardhan and Thomas Degry and Thomas Shadwell and Tianfu Fu and Tianshi Gao and Timur Garipov and Tina Sriskandarajah and Toki Sherbakov and Tomer Kaftan and Tomo Hiratsuka and Tongzhou Wang and Tony Song and Tony Zhao and Troy Peterson and Val Kharitonov and Victoria Chernova and Vineet Kosaraju and Vishal Kuo and Vitchyr Pong and Vivek Verma and Vlad Petrov and Wanning Jiang and Weixing Zhang and Wenda Zhou and Wenlei Xie and Wenting Zhan and Wes McCabe and Will DePue and Will Ellsworth and Wulfie Bain and Wyatt Thompson and Xiangning Chen and Xiangyu Qi and Xin Xiang and Xinwei Shi and Yann Dubois and Yaodong Yu and Yara Khakbaz and Yifan Wu and Yilei Qian and Yin Tat Lee and Yinbo Chen and Yizhen Zhang and Yizhong Xiong and Yonglong Tian and Young Cha and Yu Bai and Yu Yang and Yuan Yuan and Yuanzhi Li and Yufeng Zhang and Yuguang Yang and Yujia Jin and Yun Jiang and Yunyun Wang and Yushi Wang and Yutian Liu and Zach Stubenvoll and Zehao Dou and Zheng Wu and Zhigang Wang},
      year={2025},
      eprint={2601.03267},
      archivePrefix={arXiv},
      primaryClass={cs.CL},
      url={https://arxiv.org/abs/2601.03267}, 
}

@misc{lu2024mathvistaevaluatingmathematicalreasoning,
      title={MathVista: Evaluating Mathematical Reasoning of Foundation Models in Visual Contexts}, 
      author={Pan Lu and Hritik Bansal and Tony Xia and Jiacheng Liu and Chunyuan Li and Hannaneh Hajishirzi and Hao Cheng and Kai-Wei Chang and Michel Galley and Jianfeng Gao},
      year={2024},
      eprint={2310.02255},
      archivePrefix={arXiv},
      primaryClass={cs.CV},
      url={https://arxiv.org/abs/2310.02255}, 
}

@inproceedings{sun-etal-2024-mm,
    title = "{MM}-{MATH}: Advancing Multimodal Math Evaluation with Process Evaluation and Fine-grained Classification",
    author = "Sun, Kai  and
      Bai, Yushi  and
      Qi, Ji  and
      Hou, Lei  and
      Li, Juanzi",
    editor = "Al-Onaizan, Yaser  and
      Bansal, Mohit  and
      Chen, Yun-Nung",
    booktitle = "Findings of the Association for Computational Linguistics: EMNLP 2024",
    month = nov,
    year = "2024",
    address = "Miami, Florida, USA",
    publisher = "Association for Computational Linguistics",
    url = "https://aclanthology.org/2024.findings-emnlp.73/",
    doi = "10.18653/v1/2024.findings-emnlp.73",
    pages = "1358--1375"
}

@misc{xiang2025seephysdoesseeinghelp,
      title={SeePhys: Does Seeing Help Thinking? -- Benchmarking Vision-Based Physics Reasoning}, 
      author={Kun Xiang and Heng Li and Terry Jingchen Zhang and Yinya Huang and Zirong Liu and Peixin Qu and Jixi He and Jiaqi Chen and Yu-Jie Yuan and Jianhua Han and Hang Xu and Hanhui Li and Mrinmaya Sachan and Xiaodan Liang},
      year={2025},
      eprint={2505.19099},
      archivePrefix={arXiv},
      primaryClass={cs.AI},
      url={https://arxiv.org/abs/2505.19099}, 
}

@misc{qiu2025phybenchholisticevaluationphysical,
      title={PHYBench: Holistic Evaluation of Physical Perception and Reasoning in Large Language Models}, 
      author={Shi Qiu and Shaoyang Guo and Zhuo-Yang Song and Yunbo Sun and Zeyu Cai and Jiashen Wei and Tianyu Luo and Yixuan Yin and Haoxu Zhang and Yi Hu and Chenyang Wang and Chencheng Tang and Haoling Chang and Qi Liu and Ziheng Zhou and Tianyu Zhang and Jingtian Zhang and Zhangyi Liu and Minghao Li and Yuku Zhang and Boxuan Jing and Xianqi Yin and Yutong Ren and Zizhuo Fu and Jiaming Ji and Weike Wang and Xudong Tian and Anqi Lv and Laifu Man and Jianxiang Li and Feiyu Tao and Qihua Sun and Zhou Liang and Yushu Mu and Zhongxuan Li and Jing-Jun Zhang and Shutao Zhang and Xiaotian Li and Xingqi Xia and Jiawei Lin and Zheyu Shen and Jiahang Chen and Qiuhao Xiong and Binran Wang and Fengyuan Wang and Ziyang Ni and Bohan Zhang and Fan Cui and Changkun Shao and Qing-Hong Cao and Ming-xing Luo and Yaodong Yang and Muhan Zhang and Hua Xing Zhu},
      year={2025},
      eprint={2504.16074},
      archivePrefix={arXiv},
      primaryClass={cs.CL},
      url={https://arxiv.org/abs/2504.16074}, 
}

@misc{zhang2025physreasoncomprehensivebenchmarkphysicsbased,
      title={PhysReason: A Comprehensive Benchmark towards Physics-Based Reasoning}, 
      author={Xinyu Zhang and Yuxuan Dong and Yanrui Wu and Jiaxing Huang and Chengyou Jia and Basura Fernando and Mike Zheng Shou and Lingling Zhang and Jun Liu},
      year={2025},
      eprint={2502.12054},
      archivePrefix={arXiv},
      primaryClass={cs.AI},
      url={https://arxiv.org/abs/2502.12054}, 
}

@misc{lu2022learnexplainmultimodalreasoning,
      title={Learn to Explain: Multimodal Reasoning via Thought Chains for Science Question Answering}, 
      author={Pan Lu and Swaroop Mishra and Tony Xia and Liang Qiu and Kai-Wei Chang and Song-Chun Zhu and Oyvind Tafjord and Peter Clark and Ashwin Kalyan},
      year={2022},
      eprint={2209.09513},
      archivePrefix={arXiv},
      primaryClass={cs.CL},
      url={https://arxiv.org/abs/2209.09513}, 
}

@misc{guo2025sciverseunveilingknowledgecomprehension,
      title={SciVerse: Unveiling the Knowledge Comprehension and Visual Reasoning of LMMs on Multi-modal Scientific Problems}, 
      author={Ziyu Guo and Ray Zhang and Hao Chen and Jialin Gao and Dongzhi Jiang and Jiaze Wang and Pheng-Ann Heng},
      year={2025},
      eprint={2503.10627},
      archivePrefix={arXiv},
      primaryClass={cs.CV},
      url={https://arxiv.org/abs/2503.10627}, 
}

@misc{zheng2024largelanguagemodelsrobust,
      title={Large Language Models Are Not Robust Multiple Choice Selectors}, 
      author={Chujie Zheng and Hao Zhou and Fandong Meng and Jie Zhou and Minlie Huang},
      year={2024},
      eprint={2309.03882},
      archivePrefix={arXiv},
      primaryClass={cs.CL},
      url={https://arxiv.org/abs/2309.03882}, 
}

@misc{gull2026engtracesymbolicbenchmarkverifiable,
      title={EngTrace: A Symbolic Benchmark for Verifiable Process Supervision of Engineering Reasoning}, 
      author={Ayesha Gull and Muhammad Usman Safder and Rania Elbadry and Fan Zhang and Veselin Stoyanov and Preslav Nakov and Zhuohan Xie},
      year={2026},
      eprint={2511.01650},
      archivePrefix={arXiv},
      primaryClass={cs.CL},
      url={https://arxiv.org/abs/2511.01650}, 
}

@misc{zong2024foolvisionandlanguage,
      title={Fool Your (Vision and) Language Model With Embarrassingly Simple Permutations}, 
      author={Yongshuo Zong and Tingyang Yu and Ruchika Chavhan and Bingchen Zhao and Timothy Hospedales},
      year={2024},
      eprint={2310.01651},
      archivePrefix={arXiv},
      primaryClass={cs.LG},
      url={https://arxiv.org/abs/2310.01651}, 
}

@misc{mirzadeh2025gsmsymbolicunderstandinglimitationsmathematical,
      title={GSM-Symbolic: Understanding the Limitations of Mathematical Reasoning in Large Language Models}, 
      author={Iman Mirzadeh and Keivan Alizadeh and Hooman Shahrokhi and Oncel Tuzel and Samy Bengio and Mehrdad Farajtabar},
      year={2025},
      eprint={2410.05229},
      archivePrefix={arXiv},
      primaryClass={cs.LG},
      url={https://arxiv.org/abs/2410.05229}, 
}

@misc{zou2025dynamathdynamicvisualbenchmark,
      title={DynaMath: A Dynamic Visual Benchmark for Evaluating Mathematical Reasoning Robustness of Vision Language Models}, 
      author={Chengke Zou and Xingang Guo and Rui Yang and Junyu Zhang and Bin Hu and Huan Zhang},
      year={2025},
      eprint={2411.00836},
      archivePrefix={arXiv},
      primaryClass={cs.CV},
      url={https://arxiv.org/abs/2411.00836}, 
}

@misc{ren2025vgrpbenchvisualgridreasoning,
      title={VGRP-Bench: Visual Grid Reasoning Puzzle Benchmark for Large Vision-Language Models}, 
      author={Yufan Ren and Konstantinos Tertikas and Shalini Maiti and Junlin Han and Tong Zhang and Sabine Süsstrunk and Filippos Kokkinos},
      year={2025},
      eprint={2503.23064},
      archivePrefix={arXiv},
      primaryClass={cs.CV},
      url={https://arxiv.org/abs/2503.23064}, 
}

@misc{shinoda2025agrobenchvisionlanguagemodelbenchmark,
      title={AgroBench: Vision-Language Model Benchmark in Agriculture}, 
      author={Risa Shinoda and Nakamasa Inoue and Hirokatsu Kataoka and Masaki Onishi and Yoshitaka Ushiku},
      year={2025},
      eprint={2507.20519},
      archivePrefix={arXiv},
      primaryClass={cs.CV},
      url={https://arxiv.org/abs/2507.20519}, 
}

@inproceedings{nayak-etal-2024-benchmarking,
    title = "Benchmarking Vision Language Models for Cultural Understanding",
    author = "Nayak, Shravan  and
      Jain, Kanishk  and
      Awal, Rabiul  and
      Reddy, Siva  and
      Steenkiste, Sjoerd Van  and
      Hendricks, Lisa Anne  and
      Stanczak, Karolina  and
      Agrawal, Aishwarya",
    editor = "Al-Onaizan, Yaser  and
      Bansal, Mohit  and
      Chen, Yun-Nung",
    booktitle = "Proceedings of the 2024 Conference on Empirical Methods in Natural Language Processing",
    month = nov,
    year = "2024",
    address = "Miami, Florida, USA",
    publisher = "Association for Computational Linguistics",
    url = "https://aclanthology.org/2024.emnlp-main.329/",
    doi = "10.18653/v1/2024.emnlp-main.329",
    pages = "5769--5790"
}

@inproceedings{nyandwi-etal-2025-grounding,
    title = "Grounding Multilingual Multimodal {LLM}s With Cultural Knowledge",
    author = "Nyandwi, Jean De Dieu  and
      Song, Yueqi  and
      Khanuja, Simran  and
      Neubig, Graham",
    editor = "Christodoulopoulos, Christos  and
      Chakraborty, Tanmoy  and
      Rose, Carolyn  and
      Peng, Violet",
    booktitle = "Proceedings of the 2025 Conference on Empirical Methods in Natural Language Processing",
    month = nov,
    year = "2025",
    address = "Suzhou, China",
    publisher = "Association for Computational Linguistics",
    url = "https://aclanthology.org/2025.emnlp-main.1232/",
    doi = "10.18653/v1/2025.emnlp-main.1232",
    pages = "24198--24242",
    ISBN = "979-8-89176-332-6"
}

@misc{deepmind2025gemini25pro,
  author       = {{Google DeepMind}},
  title        = {Gemini 2.5 Pro Model Card},
  year         = {2025},
  howpublished = {\url{https://storage.googleapis.com/deepmind-media/Model-Cards/Gemini-2-5-Pro-Model-Card.pdf}},
  publisher    = {Google},
  note         = {Accessed: 2026-04-26}
}

@misc{deepmind2025gemini25flash,
  author       = {{Google DeepMind}},
  title        = {Gemini 2.5 Flash Model Card},
  year         = {2025},
  howpublished = {\url{https://storage.googleapis.com/deepmind-media/Model-Cards/Gemini-2-5-Flash-Model-Card.pdf}},
  publisher    = {Google},
  note         = {Accessed: 2026-04-26}
}

@misc{gemmateam2025gemma3technicalreport,
      title={Gemma 3 Technical Report}, 
      author={Gemma Team and Aishwarya Kamath and Johan Ferret and Shreya Pathak and Nino Vieillard and Ramona Merhej and Sarah Perrin and Tatiana Matejovicova and Alexandre Ramé and Morgane Rivière and Louis Rouillard and Thomas Mesnard and Geoffrey Cideron and Jean-bastien Grill and Sabela Ramos and Edouard Yvinec and Michelle Casbon and Etienne Pot and Ivo Penchev and Gaël Liu and Francesco Visin and Kathleen Kenealy and Lucas Beyer and Xiaohai Zhai and Anton Tsitsulin and Robert Busa-Fekete and Alex Feng and Noveen Sachdeva and Benjamin Coleman and Yi Gao and Basil Mustafa and Iain Barr and Emilio Parisotto and David Tian and Matan Eyal and Colin Cherry and Jan-Thorsten Peter and Danila Sinopalnikov and Surya Bhupatiraju and Rishabh Agarwal and Mehran Kazemi and Dan Malkin and Ravin Kumar and David Vilar and Idan Brusilovsky and Jiaming Luo and Andreas Steiner and Abe Friesen and Abhanshu Sharma and Abheesht Sharma and Adi Mayrav Gilady and Adrian Goedeckemeyer and Alaa Saade and Alex Feng and Alexander Kolesnikov and Alexei Bendebury and Alvin Abdagic and Amit Vadi and András György and André Susano Pinto and Anil Das and Ankur Bapna and Antoine Miech and Antoine Yang and Antonia Paterson and Ashish Shenoy and Ayan Chakrabarti and Bilal Piot and Bo Wu and Bobak Shahriari and Bryce Petrini and Charlie Chen and Charline Le Lan and Christopher A. Choquette-Choo and CJ Carey and Cormac Brick and Daniel Deutsch and Danielle Eisenbud and Dee Cattle and Derek Cheng and Dimitris Paparas and Divyashree Shivakumar Sreepathihalli and Doug Reid and Dustin Tran and Dustin Zelle and Eric Noland and Erwin Huizenga and Eugene Kharitonov and Frederick Liu and Gagik Amirkhanyan and Glenn Cameron and Hadi Hashemi and Hanna Klimczak-Plucińska and Harman Singh and Harsh Mehta and Harshal Tushar Lehri and Hussein Hazimeh and Ian Ballantyne and Idan Szpektor and Ivan Nardini and Jean Pouget-Abadie and Jetha Chan and Joe Stanton and John Wieting and Jonathan Lai and Jordi Orbay and Joseph Fernandez and Josh Newlan and Ju-yeong Ji and Jyotinder Singh and Kat Black and Kathy Yu and Kevin Hui and Kiran Vodrahalli and Klaus Greff and Linhai Qiu and Marcella Valentine and Marina Coelho and Marvin Ritter and Matt Hoffman and Matthew Watson and Mayank Chaturvedi and Michael Moynihan and Min Ma and Nabila Babar and Natasha Noy and Nathan Byrd and Nick Roy and Nikola Momchev and Nilay Chauhan and Noveen Sachdeva and Oskar Bunyan and Pankil Botarda and Paul Caron and Paul Kishan Rubenstein and Phil Culliton and Philipp Schmid and Pier Giuseppe Sessa and Pingmei Xu and Piotr Stanczyk and Pouya Tafti and Rakesh Shivanna and Renjie Wu and Renke Pan and Reza Rokni and Rob Willoughby and Rohith Vallu and Ryan Mullins and Sammy Jerome and Sara Smoot and Sertan Girgin and Shariq Iqbal and Shashir Reddy and Shruti Sheth and Siim Põder and Sijal Bhatnagar and Sindhu Raghuram Panyam and Sivan Eiger and Susan Zhang and Tianqi Liu and Trevor Yacovone and Tyler Liechty and Uday Kalra and Utku Evci and Vedant Misra and Vincent Roseberry and Vlad Feinberg and Vlad Kolesnikov and Woohyun Han and Woosuk Kwon and Xi Chen and Yinlam Chow and Yuvein Zhu and Zichuan Wei and Zoltan Egyed and Victor Cotruta and Minh Giang and Phoebe Kirk and Anand Rao and Kat Black and Nabila Babar and Jessica Lo and Erica Moreira and Luiz Gustavo Martins and Omar Sanseviero and Lucas Gonzalez and Zach Gleicher and Tris Warkentin and Vahab Mirrokni and Evan Senter and Eli Collins and Joelle Barral and Zoubin Ghahramani and Raia Hadsell and Yossi Matias and D. Sculley and Slav Petrov and Noah Fiedel and Noam Shazeer and Oriol Vinyals and Jeff Dean and Demis Hassabis and Koray Kavukcuoglu and Clement Farabet and Elena Buchatskaya and Jean-Baptiste Alayrac and Rohan Anil and Dmitry and Lepikhin and Sebastian Borgeaud and Olivier Bachem and Armand Joulin and Alek Andreev and Cassidy Hardin and Robert Dadashi and Léonard Hussenot},
      year={2025},
      eprint={2503.19786},
      archivePrefix={arXiv},
      primaryClass={cs.CL},
      url={https://arxiv.org/abs/2503.19786}, 
}

@misc{wang2025internvl35advancingopensourcemultimodal,
      title={InternVL3.5: Advancing Open-Source Multimodal Models in Versatility, Reasoning, and Efficiency}, 
      author={Weiyun Wang and Zhangwei Gao and Lixin Gu and Hengjun Pu and Long Cui and Xingguang Wei and Zhaoyang Liu and Linglin Jing and Shenglong Ye and Jie Shao and Zhaokai Wang and Zhe Chen and Hongjie Zhang and Ganlin Yang and Haomin Wang and Qi Wei and Jinhui Yin and Wenhao Li and Erfei Cui and Guanzhou Chen and Zichen Ding and Changyao Tian and Zhenyu Wu and Jingjing Xie and Zehao Li and Bowen Yang and Yuchen Duan and Xuehui Wang and Zhi Hou and Haoran Hao and Tianyi Zhang and Songze Li and Xiangyu Zhao and Haodong Duan and Nianchen Deng and Bin Fu and Yinan He and Yi Wang and Conghui He and Botian Shi and Junjun He and Yingtong Xiong and Han Lv and Lijun Wu and Wenqi Shao and Kaipeng Zhang and Huipeng Deng and Biqing Qi and Jiaye Ge and Qipeng Guo and Wenwei Zhang and Songyang Zhang and Maosong Cao and Junyao Lin and Kexian Tang and Jianfei Gao and Haian Huang and Yuzhe Gu and Chengqi Lyu and Huanze Tang and Rui Wang and Haijun Lv and Wanli Ouyang and Limin Wang and Min Dou and Xizhou Zhu and Tong Lu and Dahua Lin and Jifeng Dai and Weijie Su and Bowen Zhou and Kai Chen and Yu Qiao and Wenhai Wang and Gen Luo},
      year={2025},
      eprint={2508.18265},
      archivePrefix={arXiv},
      primaryClass={cs.CV},
      url={https://arxiv.org/abs/2508.18265}, 
}

@misc{clark2026molmo2openweightsdata,
      title={Molmo2: Open Weights and Data for Vision-Language Models with Video Understanding and Grounding}, 
      author={Christopher Clark and Jieyu Zhang and Zixian Ma and Jae Sung Park and Mohammadreza Salehi and Rohun Tripathi and Sangho Lee and Zhongzheng Ren and Chris Dongjoo Kim and Yinuo Yang and Vincent Shao and Yue Yang and Weikai Huang and Ziqi Gao and Taira Anderson and Jianrui Zhang and Jitesh Jain and George Stoica and Winson Han and Ali Farhadi and Ranjay Krishna},
      year={2026},
      eprint={2601.10611},
      archivePrefix={arXiv},
      primaryClass={cs.CV},
      url={https://arxiv.org/abs/2601.10611}, 
}

@misc{aisingapore2026sealionv4,
  author       = {{AI Singapore}},
  title        = {SEA-LION v4 Model Documentation},
  year         = {2026},
  howpublished = {\url{https://docs.sea-lion.ai/models/sea-lion-v4}},
  publisher    = {AI Singapore},
  note         = {Accessed: 2026-04-26}
}

@misc{bai2025qwen3vltechnicalreport,
      title={Qwen3-VL Technical Report}, 
      author={Shuai Bai and Yuxuan Cai and Ruizhe Chen and Keqin Chen and Xionghui Chen and Zesen Cheng and Lianghao Deng and Wei Ding and Chang Gao and Chunjiang Ge and Wenbin Ge and Zhifang Guo and Qidong Huang and Jie Huang and Fei Huang and Binyuan Hui and Shutong Jiang and Zhaohai Li and Mingsheng Li and Mei Li and Kaixin Li and Zicheng Lin and Junyang Lin and Xuejing Liu and Jiawei Liu and Chenglong Liu and Yang Liu and Dayiheng Liu and Shixuan Liu and Dunjie Lu and Ruilin Luo and Chenxu Lv and Rui Men and Lingchen Meng and Xuancheng Ren and Xingzhang Ren and Sibo Song and Yuchong Sun and Jun Tang and Jianhong Tu and Jianqiang Wan and Peng Wang and Pengfei Wang and Qiuyue Wang and Yuxuan Wang and Tianbao Xie and Yiheng Xu and Haiyang Xu and Jin Xu and Zhibo Yang and Mingkun Yang and Jianxin Yang and An Yang and Bowen Yu and Fei Zhang and Hang Zhang and Xi Zhang and Bo Zheng and Humen Zhong and Jingren Zhou and Fan Zhou and Jing Zhou and Yuanzhi Zhu and Ke Zhu},
      year={2025},
      eprint={2511.21631},
      archivePrefix={arXiv},
      primaryClass={cs.CV},
      url={https://arxiv.org/abs/2511.21631}, 
}

@misc{chen2025truemultimodalincontextlearning,
      title={True Multimodal In-Context Learning Needs Attention to the Visual Context}, 
      author={Shuo Chen and Jianzhe Liu and Zhen Han and Yan Xia and Daniel Cremers and Philip Torr and Volker Tresp and Jindong Gu},
      year={2025},
      eprint={2507.15807},
      archivePrefix={arXiv},
      primaryClass={cs.CV},
      url={https://arxiv.org/abs/2507.15807}, 
}

@misc{deng2025wordsvisionvisionlanguagemodels,
      title={Words or Vision: Do Vision-Language Models Have Blind Faith in Text?}, 
      author={Ailin Deng and Tri Cao and Zhirui Chen and Bryan Hooi},
      year={2025},
      eprint={2503.02199},
      archivePrefix={arXiv},
      primaryClass={cs.CV},
      url={https://arxiv.org/abs/2503.02199}, 
}

@misc{yin2025refusalfallscliffsafety,
      title={Refusal Falls off a Cliff: How Safety Alignment Fails in Reasoning?}, 
      author={Qingyu Yin and Chak Tou Leong and Linyi Yang and Wenxuan Huang and Wenjie Li and Xiting Wang and Jaehong Yoon and YunXing and XingYu and Jinjin Gu},
      year={2025},
      eprint={2510.06036},
      archivePrefix={arXiv},
      primaryClass={cs.AI},
      url={https://arxiv.org/abs/2510.06036}, 
}

@misc{zheng2025mllmsdeeplyaffectedmodality,
      title={MLLMs are Deeply Affected by Modality Bias}, 
      author={Xu Zheng and Chenfei Liao and Yuqian Fu and Kaiyu Lei and Yuanhuiyi Lyu and Lutao Jiang and Bin Ren and Jialei Chen and Jiawen Wang and Chengxin Li and Linfeng Zhang and Danda Pani Paudel and Xuanjing Huang and Yu-Gang Jiang and Nicu Sebe and Dacheng Tao and Luc Van Gool and Xuming Hu},
      year={2025},
      eprint={2505.18657},
      archivePrefix={arXiv},
      primaryClass={cs.AI},
      url={https://arxiv.org/abs/2505.18657}, 
}
\bibliographystyle{iclr2027_conference}

\appendix
\section{Additional Information for Dataset Construction} \label{appendix:a}

\subsection{Curriculum and Taxonomy} \label{appendix:a0}

\paragraph{Cambridge IGCSE and A-Level}
The Cambridge International General Certificate of Secondary Education (IGCSE) and the Cambridge Advanced Level (A-Level) are international secondary education qualifications developed by Cambridge Assessment International Education. IGCSE is a foundational secondary curriculum typically taken by students aged 14--16, while A-Level is an advanced pre-university curriculum for students aged 16--18. Both curricula define a standardized scope of high school STEM competencies, which we adopt as the basis of our topic hierarchy.

\paragraph{Vision-Dependent Materials}
We define vision-dependent materials as subtopics in which text alone is insufficient to describe or solve a problem, so that the image carries information required to reach the answer. Examples include electric circuit analysis, graph interpretation (e.g., finding the tangent of a parabola), dimensions of geometric shapes, and molecular structures. Subtopics such as pure algebraic equation solving or text-only word problems are less vision-dependent, and we did not include them in our dataset.

\paragraph{Topics and Subtopics}
Our taxonomy has two levels. A topic contains one or more subtopics, and each subtopic belongs to exactly one topic. Each template is assigned to exactly one subtopic, and therefore to exactly one topic.

\subsection{Template Creation} \label{appendix:a1}
The dataset was constructed through a collaboration between the authors and hired contributors. All individuals involved were proficient in Python and the corresponding STEM subject, with backgrounds as STEM graduates, advanced undergraduates, or science olympiad medalists. Their responsibilities included understanding the vision-dependent subtopics of the corresponding subject, developing dynamic problem templates (complete with images, descriptions, solutions, and reasoning steps), and cross-validating each other's work. The topics and subtopics were organized beforehand by the authors, predominantly based on the Cambridge IGCSE and A-Level curricula (Appendix~\ref{appendix:a0}). In total, eight people (including authors and non-authors) participated in the construction of the \datasetsize{} templates. All hired contributors (non-authors) were compensated at a rate above the minimum wage of their region of residence.

The template creation process began with a one-hour workshop, in which we explained the research objectives and workflows, including a live demonstration of creating a template. Following this, the contributors were given five weeks (25 working days) to develop the code, with a maximum limit of 100 templates per person. Table~\ref{tab:template_guideline} summarizes the written guideline given to the contributors. Some details shown during the workshop demonstration may not be fully captured in the written guideline. Contributors could adapt problems from existing sources, create entirely new problems, or combine both. When adapting a problem, they were required to record its source in the dataset.

\begin{table}[htbp]
    \caption{Guideline given to contributors for template creation.}
    \centering
    \begin{tcolorbox}[
        colback=white,
        colframe=darkgray,
        coltitle=white,
        title=Guideline -- Template Creation,
        fonttitle=\bfseries,
        fontupper=\small,
        arc=3mm,
        boxrule=1.5pt,
        left=2mm,
        right=2mm,
        top=1mm,
        bottom=1mm,
        width=1\textwidth
    ]

    A \textbf{template} is the unit of problem in the dataset, and is represented by a Python class.

    \vspace{0.5em}
    \noindent \textbf{Steps:}
    \begin{enumerate}
        \setlength\itemsep{0em}
        \item Open the syllabus of the corresponding subject and curriculum.
        \item Read and understand the competency standards of a subtopic, including its scope and limitations.
        \item Write the template problem for the subtopic. Make sure the problem fits the subtopic requirements.
        \item Write the solution steps in LaTeX with the help of an LLM (e.g., GPT-5 or Gemini), using the provided prompt template (Table~\ref{tab:prompt_template_latex}).
        \item Run \texttt{python generate\_dataset.py -{}-subject physics -{}-debug True} to generate the instances.
        \item Open \texttt{dataset/physics/dataset.json} and the corresponding images to verify that the instances were generated successfully.
        \item Open \texttt{dataset/physics/debug/[file].txt} to check that the LaTeX solution is written correctly (e.g., by rendering it with QuickLaTeX).
    \end{enumerate}

    \vspace{0.5em}
    \noindent \textbf{Note:} You do not have to come up with question ideas from scratch. You may search for questions on similar materials online (e.g., previous benchmarks) and use them as inspiration. However, every question must have a clear source, recorded in the \texttt{source} field of the dataset.

    \end{tcolorbox}
    \label{tab:template_guideline}
\end{table}

\paragraph{Variable Sampling}
All randomly generated variables in a template (e.g., \texttt{self.time}) are instantiated with Python's \texttt{random} module using a predetermined seed, which makes every instance reproducible. The sampling function depends on the needs of the problem: \texttt{random.randint} for integers, \texttt{random.choice} for selecting from a list of candidates (e.g., function types or colors), and \texttt{random.uniform} for floating-point values. We set the range of each variable to resemble the values found in typical STEM exams, and added constraints so that no sampled value is physically impossible. For instance, the number of atoms must be an integer, and the speed of an object cannot exceed the speed of light.

\paragraph{Image Generation}
Almost all images are drawn programmatically using graphics libraries such as \texttt{matplotlib} or \texttt{PIL}. When an instance is generated, the code reads the sampled variables (e.g., \texttt{self.voltage}) and draws the image from scratch, so that the text labels, curves, shapes, and object counts in the image always match the description. A small number of templates start from an existing base diagram in PNG format. For these templates, the values displayed on the original diagram are overwritten with the newly sampled values by overlaying text at fixed coordinates using \texttt{matplotlib.pyplot.text}.

\paragraph{Solution and Verification}
The contributors wrote the reasoning steps of each template in LaTeX format. They were allowed to use AI to convert the Python computation into LaTeX (the prompt template is shown in Table~\ref{tab:prompt_template_latex}). After that, the contributors generated instances and ensured that all information was generated correctly. Contributors were required to revise the code if any component failed to generate properly. We also used an automated script to verify that each template could generate at least 10 variants, where each variant differs from the others in at least the description or the image.

For quality checking, the contributors cross-validated each other's work if at least two people worked on the same subject. When only a single contributor was assigned to a subject, a contributor from a different subject reviewed their work. Reviewers verified the correctness of each template, including the problem category, description, answer, image, and reasoning, and generated instances to make sure the template was written correctly.


\subsection{Translation}\label{appendix:a2}
The dataset was translated into six languages (Indonesian, Chinese, Arabic, Hindi, Kazakh, and Swahili) to evaluate VLM performance in non-English contexts. First, all problem descriptions from the Python templates were consolidated into a single spreadsheet. We then used Gemini 3 Pro to generate initial translation drafts, using the prompt template in Table~\ref{tab:prompt_template_translation}. We selected this model for its proficiency in multilingual contexts and its ability to preserve Python syntax.

The translation and verification process involved a collaboration between authors and hired contributors. To ensure consistency with the original dataset, we prioritized contributions from authors and workers already involved in the English dataset development. When no native speaker of a target language was present among the primary contributors, we hired external native-speaking translators for that language. All individuals involved in the translation verification were native speakers of the target language, highly familiar with A-Level STEM terminology, and proficient in Python.

All translators were given a guideline document to ensure consistency, summarized in Table~\ref{tab:translation_guideline}. The guideline asks translators to review the machine-translated draft critically against the English original, handle scientific terminology according to the conventions of the target language, and keep the Python and LaTeX syntax intact so that every translated description still runs inside its template. The verification of \datasetsize{} descriptions was completed over a three-week period. To maintain quality control, we intentionally introduced errors into the dataset by deleting the middle 25\% of 40 randomly chosen descriptions (approximately 7\% of the total). Translators who failed to identify and correct these errors were required to revise their work, as this served as a proxy for attention to detail. Upon successful completion of the task, all annotators were compensated at competitive rates.

\begin{table}[htbp]
    \caption{Summary of the guideline given to translators for verifying machine-translated descriptions.}
    \centering
    \begin{tcolorbox}[
        colback=white,
        colframe=darkgray,
        coltitle=white,
        title=Guideline -- Translation Verification,
        fonttitle=\bfseries,
        fontupper=\small,
        arc=3mm,
        boxrule=1.5pt,
        left=2mm,
        right=2mm,
        top=1mm,
        bottom=1mm,
        width=1\textwidth
    ]

    Your task is to review and refine AI-generated translations of STEM problems. The final text must be scientifically accurate, natural in the target language, and correctly formatted for Python code integration.

    \vspace{0.5em}
    \noindent \textbf{Step 1 -- Compare Source and Draft}
    \begin{itemize}
        \setlength\itemsep{0em}
        \item Read the English original first to understand the discipline and context (e.g., kinematics, calculus, organic chemistry, or genetics).
        \item Do not assume the AI draft is correct. Look for awkward phrasing, literal translations of idioms, and misunderstandings of the underlying science.
    \end{itemize}

    \noindent \textbf{Step 2 -- Scientific Terminology}
    \begin{itemize}
        \setlength\itemsep{0em}
        \item Follow the educational standard of the target language for acronyms (e.g., DNA becomes ADN in Spanish). Keep English letters if the local curriculum does.
        \item Use established localized terms where they exist (e.g., ``velocity'' becomes ``kecepatan'' in Indonesian). Keep a term in English if no widely used equivalent exists.
        \item Never translate or transliterate Latin binomial names (e.g., \textit{Escherichia coli}). Translate the common name if given.
        \item Translate IUPAC chemical names (e.g., sodium chloride) into the formal chemical nomenclature of the target language.
        \item Preserve the letters of variables and constants (e.g., do not change $F$ for Force into $K$ for \textit{Kraft} in German).
    \end{itemize}

    \noindent \textbf{Step 3 -- Protect Python Syntax}
    \begin{itemize}
        \setlength\itemsep{0em}
        \item Do not translate, re-case, or remove anything inside \texttt{\{ \}}. The block may be moved within the sentence if grammar requires it.
        \item Keep double backslashes and escape characters (e.g., \texttt{\textbackslash\textbackslash frac}, \texttt{\textbackslash n}) exactly as in the original.
        \item Respect string boundaries. Escape any quote character that matches the outer wrapper.
        \item Do not insert hard line breaks unless the original uses triple quotes or explicit \texttt{\textbackslash n}.
        \item Localize decimal separators in plain text only. Never change format specifiers such as \texttt{\{self.speed:.2f\}}.
        \item Keep string prefixes (\texttt{f}, \texttt{r}, \texttt{rf}) as they are.
    \end{itemize}

    \noindent \textbf{Step 4 -- Mathematical Symbols and Units}
    \begin{itemize}
        \setlength\itemsep{0em}
        \item Keep scientific and mathematical notation exactly as it is.
        \item For right-to-left languages (e.g., Arabic), make sure formulas and chemical reactions still read left to right.
        \item Preserve the \texttt{\$} math delimiters. Do not place translated words inside \texttt{\$...\$} or \texttt{\textbackslash text\{\}}, since the math font used for rendering cannot display non-Latin characters.
    \end{itemize}

    \noindent \textbf{Step 5 -- Natural Phrasing}
    \begin{itemize}
        \setlength\itemsep{0em}
        \item Rewrite robotic sentences so they read naturally to a native speaker, in the formal register of a STEM textbook or exam.
        \item Use the standard localized form of units where that is the convention in the target language.
    \end{itemize}

    \noindent \textbf{Additional Notes}
    \begin{itemize}
        \setlength\itemsep{0em}
        \item Translate references to visual elements (e.g., ``the shaded region'', ``point A'') literally, without re-wording elements you cannot see.
        \item Only modify the column containing the translated text.
    \end{itemize}

    \end{tcolorbox}
    \label{tab:translation_guideline}
\end{table}

\begin{table}[hp]
    \caption{Prompt template for converting Python reasoning to LaTeX format.}
    
    \centering
    
    \begin{tcolorbox}[
        colback=white,
        colframe=darkgray,
        coltitle=white,
        title=Prompt Template -- Generating LaTeX,
        fonttitle=\bfseries,
        fontupper=\small,
        arc=3mm,
        boxrule=1.5pt,
        left=2mm,
        right=2mm,
        top=1mm,
        bottom=1mm,
        width=1\textwidth 
    ]
    
    As an expert in Python and LaTeX formatting, your task is to convert the reasoning process of my Python code into a LaTeX-formatted explanation. Store the result in a single variable named \texttt{self.reasoning}.
    
    \vspace{1em}
    \noindent \textbf{Guidelines:}
    \begin{itemize}
        \setlength\itemsep{0em}
        \item Use f-strings in Python format (with double backslashes for LaTeX newlines, e.g. \texttt{\textbackslash\textbackslash}).
        \item Wrap the entire reasoning inside \texttt{\textbackslash begin\{aligned\} ... \textbackslash end\{aligned\}} and use \texttt{\textbackslash\textbackslash} for line breaks between aligned equations.
        \item Each mathematical step must appear on a new line using \texttt{\textbackslash\textbackslash}.
        \item Use clean, step-by-step mathematical reasoning with proper LaTeX formatting.
        \item Include intermediate symbolic steps before substituting numeric values.
        \item Include units if applicable (e.g. kg, m, s).
        \item Only output the final Python code (no extra text, markdown, or explanations). Do not write anything else.
    \end{itemize}
    
    \vspace{1em}
    \noindent \textbf{EXAMPLE INPUT} \\
    \texttt{self.description = "A box experiences two perpendicular forces as shown in the diagram. The magnitude of the resultant force is (\_) N"} \\
    \texttt{self.x\_direction = rnd.randint(-18, 18)} \\
    \texttt{self.y\_direction = rnd.randint(-18, 18)} \\
    \texttt{while self.x\_direction in range(-5, 6) or self.y\_direction in range(-5, 6):} \\
    \texttt{\hspace*{2em}self.x\_direction = rnd.randint(-18, 18)} \\
    \texttt{\hspace*{2em}self.y\_direction = rnd.randint(-18, 18)} \\
    \texttt{self.resultant = math.sqrt(self.x\_direction**2 + self.y\_direction**2)} \\
    \texttt{self.answer = round(self.resultant, 2)}
    
    \vspace{1em}
    \noindent \textbf{EXPECTED OUTPUT} \\
    \texttt{self.reasoning = (} \\
    \texttt{\hspace*{2em}"\textbackslash[\textbackslash begin\{aligned\}"} \\
    \texttt{\hspace*{2em}f"\textbackslash text\{\{Horizontal component\}\} \&= \{self.x\_direction\}\textbackslash,\textbackslash text\{\{N\}\}\textbackslash\textbackslash"} \\
    \texttt{\hspace*{2em}f"\textbackslash text\{\{Vertical component\}\} \&= \{self.y\_direction\}\textbackslash,\textbackslash text\{\{N\}\}\textbackslash\textbackslash"} \\
    \texttt{\hspace*{2em}f"\textbackslash text\{\{Resultant force is given by:\}\} \&\textbackslash\textbackslash"} \\
    \texttt{\hspace*{2em}f"R \&= \textbackslash sqrt\{\{x\textasciicircum 2 + y\textasciicircum 2\}\}\textbackslash\textbackslash"} \\
    \texttt{\hspace*{2em}f"\&= \textbackslash sqrt\{\{(\{self.x\_direction\})\textasciicircum 2 + (\{self.y\_direction\})\textasciicircum 2\}\}\textbackslash\textbackslash"} \\
    \texttt{\hspace*{2em}f"\&= \textbackslash sqrt\{\{\{self.x\_direction**2\} + \{self.y\_direction**2\}\}\}\textbackslash\textbackslash"} \\
    \texttt{\hspace*{2em}f"\&= \{self.answer\}\textbackslash,\textbackslash text\{\{N\}\}"} \\
    \texttt{\hspace*{2em}"\textbackslash end\{aligned\}\textbackslash]"} \\
    \texttt{)}
    
    \vspace{1em}
    \noindent Now, here is the actual Python code you need to explain: \\
    \texttt{[CODE]}
    
    \end{tcolorbox}

    \label{tab:prompt_template_latex}
\end{table}

\begin{table}[htbp]
    \caption{Prompt template for translation. During the experiment, we prompted Gemini 3 Pro with a batch of 100 line of codes per prompt.}

    \centering
    \begin{tcolorbox}[
        colback=white,
        colframe=darkgray,
        coltitle=white,
        title=Prompt Template -- Translation,
        fonttitle=\bfseries,
        fontupper=\small, 
        arc=3mm,
        boxrule=1.5pt,
        left=2mm,
        right=2mm,
        top=1mm,
        bottom=1mm,
        width=1\textwidth
    ]
    
    Translate the following Python f-strings into [LANGUAGE].
    
    \vspace{1em}
    \noindent \textbf{Please strictly follow these rules:}
    \begin{enumerate}
        \setlength\itemsep{0em}
        \item \textbf{Format:} Output the translated code vertically, with exactly one string assignment per line so it is easy to copy-paste. Do not add extra line breaks inside the strings.
        \item \textbf{Variable Name:} Use the exact variable name \texttt{self.description\_[LANGUAGE]} for every line.
        \item \textbf{Code Syntax:} Do NOT translate, change the casing, or alter any variables inside the curly brackets \texttt{\{ \}} (e.g., \texttt{\{self.sample\_size\}}).
        \item \textbf{STEM Accuracy:} Maintain formal scientific and mathematical accuracy. Keep universal abbreviations (DNA, ATP, etc.) in English.
        \item \textbf{String Formatting \& Escaping:} Mirror the exact string prefix (\texttt{f"..."} or \texttt{"..."}), the exact backslash escaping (\texttt{\textbackslash frac} vs \texttt{\textbackslash\textbackslash frac}), and other formatting from the original.
    \end{enumerate}
    
    \vspace{1em}
    \noindent Code to translate: \\
    \texttt{[CODE]}
    
    \end{tcolorbox}
    \label{tab:prompt_template_translation}
\end{table}

\section{Variants in Our Dataset} \label{appendix:b}
\begin{figure}[hp]
    \centering
    \includegraphics[width=1.0\textwidth]{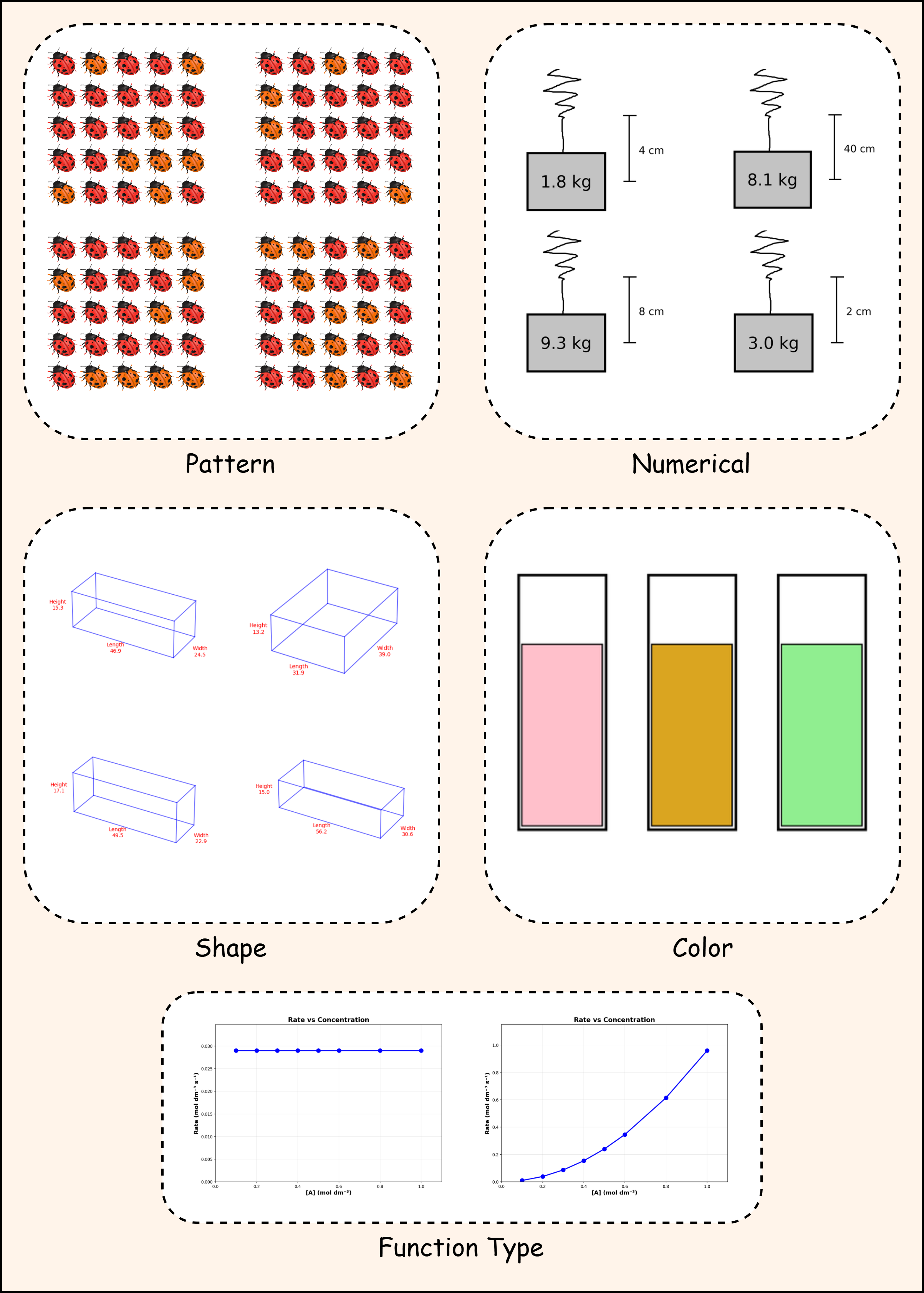}
    \caption{More examples of variants in our dataset.}
    \label{fig:variants_sample}
\end{figure}

Figure~\ref{fig:variants_sample} presents examples of the variants in our dataset. We identify five types of variants in our dataset: pattern, numerical, shape, color, and function type.

\FloatBarrier

\section{Experiment Setup}
\label{appendix:c}

\subsection{Model Artifacts}
\label{appendix:c1}
Figure~\ref{tab:model_params} displays the model parameters used for the experiment. By default, we used cloud-based GPU to collect model responses of small open-weight models. For large open-weight models (\(\ge \)27B), we used API-based inference for convenience.

\begin{table}[htbp]
    \caption{Model parameters used in the experiments.} \label{tab:model_params}
    \centering
    \small
    \renewcommand{\arraystretch}{1.3} 
    \begin{tabular}{l p{0.65\textwidth}}
        \hline
        \textbf{Model} & \textbf{Parameters} \\
        \hline
        GPT-5.4 & model\_name=\texttt{gpt-5.4-2026-03-05} \newline temperature=1.0 \newline max\_completion\_tokens=4096 \newline reasoning\_effort="none" \\
        \hline
        Gemini-2.5-Pro & model\_name=\texttt{gemini-2.5-pro} \newline temperature=0.0 \newline max\_output\_tokens=8192 \newline thinkingBudget=2048 \\
        \hline
        Gemini-2.5-Flash & model\_name=\texttt{gemini-2.5-flash} \newline temperature=0.0 \newline max\_output\_tokens=4096 \newline thinkingBudget=0 \\
        \hline
        Qwen3.5-122B-A10B & model\_name=\texttt{Qwen/Qwen3.5-122B-A10B} \newline temperature=1.0 \newline max\_completion\_tokens=8192 \newline reasoning=False \\
        \hline
        Qwen3.5-27B & model\_name=\texttt{Qwen/Qwen3.5-27B} \newline temperature=1.0 \newline max\_completion\_tokens=8192 \newline reasoning=False \\
        \hline
        Llama-4-Scout & model\_name=\texttt{meta-llama/Llama-4-Scout-17B-16E-Instruct} \newline temperature=0.0 \newline max\_completion\_tokens=4096 \\
        \hline
        InternVL3.5-14B & model\_name=\texttt{OpenGVLab/InternVL3\_5-14B} \newline temperature=0.0 \newline max\_tokens=4096 \\
        \hline
        InternVL3.5-8B & model\_name=\texttt{OpenGVLab/InternVL3\_5-8B} \newline temperature=0.0 \newline max\_tokens=4096 \\
        \hline
        InternVL3.5-4B & model\_name=\texttt{OpenGVLab/InternVL3\_5-4B} \newline temperature=0.0 \newline max\_tokens=4096 \\
        \hline
        Qwen3-VL-8B-Instruct & model\_name=\texttt{Qwen/Qwen3-VL-8B-Instruct} \newline temperature=0.0 \newline max\_tokens=4096 \\
        \hline
        Qwen3-VL-4B-Instruct & model\_name=\texttt{Qwen/Qwen3-VL-4B-Instruct} \newline temperature=0.0 \newline max\_tokens=4096 \\
        \hline
        Molmo2-8B & model\_name=\texttt{allenai/Molmo2-8B} \newline temperature=0.0 \newline max\_tokens=4096 \\
        \hline
        Molmo2-4B & model\_name=\texttt{allenai/Molmo2-4B} \newline temperature=0.0 \newline max\_tokens=4096 \\
        \hline
        Gemma-3-12B-IT & model\_name=\texttt{google/gemma-3-12b-it} \newline temperature=0.0 \newline max\_tokens=4096 \\
        \hline
        Gemma-3-4B-IT & model\_name=\texttt{google/gemma-3-4b-it} \newline temperature=0.0 \newline max\_tokens=4096 \\
        \hline
        SEA-LION-v4-8B-VL & model\_name=\texttt{aisingapore/Qwen-SEA-LION-v4-8B-VL} \newline temperature=0.0 \newline max\_tokens=4096 \\
        \hline
        SEA-LION-v4-4B-VL & model\_name=\texttt{aisingapore/Qwen-SEA-LION-v4-4B-VL} \newline temperature=0.0 \newline max\_tokens=4096 \\
        \hline
    \end{tabular}
\end{table}

\subsection{Prompt Template}
We provide the prompt templates used in our experiments. Table~\ref{tab:prompt_template_translation} presents the translation prompt. Table~\ref{tab:prompt_template_latex} shows the template for converting code steps into \LaTeX{} steps. Table~\ref{tab:prompt_template_english} contains the instruction prompt for evaluation. Finally, the prompt templates used by the judge model for step-level reasoning evaluation (precision and recall) are given in Table \ref{tab:prompt_recall} and ~\ref{tab:prompt_precision}.

\label{appendix:c2}
\begin{table}[htb]
    \caption{Prompt template for response generation.}
    
    \centering
    \begin{tcolorbox}[
        colback=white,
        colframe=darkgray,
        coltitle=white,
        title=Prompt Template -- Response Generation,
        fonttitle=\bfseries,
        fontupper=\small, 
        arc=3mm,
        boxrule=1.5pt,
        left=2mm,
        right=2mm,
        top=1mm,
        bottom=1mm,
        width=1\textwidth
    ]
    
    Solve the above problem step-by-step. You must write your mathematical or logical reasoning strictly in English within the 'solution' field. Format your reasoning neatly into explicitly numbered steps separated by newlines (e.g., "Step 1: ...\textbackslash nStep 2: ..."). Then, extract your final answer into the 'short answer' field. Your response must be valid JSON and strictly adhere to the following format. Do not include any text outside of the JSON block.
    
    \vspace{1em}
    \texttt{\{} \\
    \texttt{\hspace*{2em}"solution": "[Step-by-step solution]",} \\
    \texttt{\hspace*{2em}"short answer": "[Final numerical value only. Do not include units]"} \\
    \texttt{\}}
    
    \end{tcolorbox}
\label{tab:prompt_template_english}
\end{table}

\begin{table}[ht]
    \caption{Prompt template for the recall judge ($\mathcal{J}_{\text{rec}}$).}
    \centering
    \begin{tcolorbox}[
        colback=white,
        colframe=darkgray,
        coltitle=white,
        title=Prompt Template -- Recall Judge,
        fonttitle=\bfseries,
        fontupper=\small,
        arc=3mm,
        boxrule=1.5pt,
        left=2mm,
        right=2mm,
        top=1mm,
        bottom=1mm,
        width=1\textwidth
    ]

    You are a STEM reasoning evaluator.

    \vspace{0.5em}
    \textbf{GROUND TRUTH SOLUTION:} \texttt{"""\{gold\_solution\}"""}

    \vspace{0.5em}
    \textbf{MODEL OUTPUT:} \texttt{"""\{model\_output\}"""}

    \vspace{0.5em}
    \textbf{PROBLEM DESCRIPTION:} \texttt{"""\{problem\_description\}"""}

    \vspace{0.5em}
    If a figure is attached, use it as additional context for judging the reasoning. Both texts contain steps labeled ``Step 1'', ``Step 2'', etc. Extract them regardless of formatting (colons, bold, newlines, indentation may vary).

    \vspace{0.5em}
    First, count the number of steps in the \textbf{GROUND TRUTH}. Call this $m$. For \textbf{EVERY} gold step 1 to $m$, decide if the model output covers it. You \textbf{MUST} include every integer from 1 to $m$ as a key --- no skipping.

    \vspace{0.5em}
    A gold step is \textbf{COVERED} if the model addresses the same operation with equivalent reasoning. Accept different notation or variable names, but \textbf{NOT} wrong values or wrong operations.

    \vspace{0.5em}
    Respond \textbf{ONLY} with this JSON, no other text:

    \vspace{0.5em}
    \texttt{\{} \\
    \texttt{\hspace*{2em}"m": <number of gold steps>,} \\
    \texttt{\hspace*{2em}"coverage": \{"1": true, "2": false, ..., "<m>": true\},} \\
    \texttt{\hspace*{2em}"rationale": \{"1": "one sentence", ..., "<m>": "one sentence"\}} \\
    \texttt{\}}

    \end{tcolorbox}

    \label{tab:prompt_recall}
\end{table}

\begin{table}[htbp]
    \caption{Prompt template for the precision judge ($\mathcal{J}_{\text{pre}}$).}
    \centering
    \begin{tcolorbox}[
        colback=white,
        colframe=darkgray,
        coltitle=white,
        title=Prompt Template -- Precision Judge,
        fonttitle=\bfseries,
        fontupper=\small,
        arc=3mm,
        boxrule=1.5pt,
        left=2mm,
        right=2mm,
        top=1mm,
        bottom=1mm,
        width=1\textwidth
    ]

    You are a STEM reasoning evaluator.

    \vspace{0.5em}
    \textbf{GROUND TRUTH SOLUTION:} \texttt{"""\{gold\_solution\}"""}

    \vspace{0.5em}
    \textbf{MODEL OUTPUT:} \texttt{"""\{model\_output\}"""}

    \vspace{0.5em}
    \textbf{PROBLEM DESCRIPTION:} \texttt{"""\{problem\_description\}"""}

    \vspace{0.5em}
    If a figure is attached, use it as additional context for judging the reasoning. Both texts contain steps labeled ``Step 1'', ``Step 2'', etc. Extract them regardless of formatting (colons, bold, newlines, indentation may vary).

    \vspace{0.5em}
    First, count the number of steps in the \textbf{MODEL OUTPUT}. Call this $n$. For \textbf{EVERY} model step 1 to $n$, decide if it is valid given the ground truth. You \textbf{MUST} include every integer from 1 to $n$ as a key --- no skipping.

    \vspace{0.5em}
    A model step is \textbf{VALID} if any of these are true: (1) it matches a ground truth step directly; (2) it is a correct sub-step of a ground truth step (splitting is fine); (3) it is a correct intermediate that logically follows.

    \vspace{0.5em}
    A model step is \textbf{INVALID} if any of these are true: (1) it restates a step already completed (padding); (2) it applies a wrong formula, wrong value, or wrong operation; (3) it introduces reasoning not grounded in the problem (hallucination).

    \vspace{0.5em}
    Respond \textbf{ONLY} with this JSON, no other text:

    \vspace{0.5em}
    \texttt{\{} \\
    \texttt{\hspace*{2em}"n": <number of model steps>,} \\
    \texttt{\hspace*{2em}"validity": \{"1": true, "2": false, ..., "<n>": true\},} \\
    \texttt{\hspace*{2em}"rationale": \{"1": "one sentence", ..., "<n>": "one sentence"\}} \\
    \texttt{\}}

    \end{tcolorbox}

    \label{tab:prompt_precision}
\end{table}

\FloatBarrier

\section{Additional Results}

This section presents fine-grained results omitted from the primary paper. Table \ref{tab:stem_vision_average_case_accuracy_language} details the average-case accuracy across all languages, while Table \ref{tab:stem_vision_worst_case_accuracy_language} provides the corresponding worst-case accuracy metrics.

\begin{table}[h]
\centering
\caption{Average-case accuracy on \datasetname{}, over all languages}
\label{tab:stem_vision_average_case_accuracy_language}
\resizebox{\textwidth}{!}{%
\begin{tabular}{lrrrrrrrr}
\hline
\textbf{Model} & \textbf{ALL} & \textbf{English} & \textbf{Arabic} & \textbf{Chinese} & \textbf{Hindi} & \textbf{Indonesian} & \textbf{Kazakh} & \textbf{Swahili} \\
\hline
\multicolumn{9}{c}{\textit{Closed-source Models}} \\
\hline
GPT-5.4 & 88.8 & 89.4 & 88.0 & 88.8 & 88.1 & 88.6 & 90.2 & 88.7 \\
Gemini-2.5-Pro & 85.8 & 86.2 & 84.8 & 85.5 & 86.7 & 85.8 & 86.1 & 85.8 \\
Gemini-2.5-Flash & 80.4 & 80.7 & 80.2 & 80.3 & 80.3 & 80.7 & 80.2 & 80.2 \\
\hline
\multicolumn{9}{l}{} \\
\multicolumn{9}{c}{\textit{Open-source Models}} \\
\hline
Qwen3.5-122B-A10B & 80.8 & 82.6 & 79.8 & 80.3 & 81.0 & 81.0 & 81.3 & 79.9 \\
Llama-4-Scout & 69.2 & 69.6 & 68.1 & 70.0 & 69.8 & 69.4 & 69.0 & 68.3 \\
InternVL3.5-14B & 55.4 & 60.0 & 57.6 & 56.0 & 57.5 & 59.1 & 54.3 & 43.5 \\
InternVL3.5-8B & 59.8 & 63.1 & 61.5 & 59.1 & 61.5 & 61.4 & 60.2 & 51.8 \\
InternVL3.5-4B & 51.6 & 54.2 & 53.6 & 54.7 & 53.9 & 53.1 & 53.0 & 38.9 \\
Qwen3-VL-8B-Instruct & 65.3 & 67.2 & 66.8 & 67.5 & 62.3 & 66.8 & 66.2 & 60.1 \\
Qwen3-VL-4B-Instruct & 26.8 & 29.9 & 21.8 & 29.0 & 26.5 & 30.0 & 22.7 & 27.8 \\
Molmo2-8B & 43.9 & 46.4 & 43.9 & 44.5 & 44.4 & 47.2 & 43.7 & 37.5 \\
Molmo2-4B & 34.0 & 38.6 & 35.6 & 38.1 & 33.6 & 34.3 & 32.9 & 25.1 \\
Gemma-3-12B-IT & 53.6 & 57.4 & 51.7 & 54.8 & 52.4 & 51.5 & 54.3 & 53.5 \\
Gemma-3-4B-IT & 36.4 & 37.3 & 36.1 & 37.0 & 35.9 & 37.2 & 36.3 & 35.1 \\
SEA-LION-v4-8B-VL & 67.7 & 70.1 & 67.3 & 68.4 & 69.5 & 68.8 & 66.4 & 63.1 \\
SEA-LION-v4-4B-VL & 44.1 & 48.2 & 41.1 & 43.3 & 43.6 & 47.2 & 43.8 & 41.2 \\
qwen-qwen3-5-27b & 82.3 & 83.6 & 81.3 & 82.8 & 81.4 & 83.2 & 81.3 & 82.3 \\
\hline
\end{tabular}%
}
\end{table}

\begin{table}[h]
\centering
\caption{Worst-case accuracy on \datasetname{} across all languages.}
\label{tab:stem_vision_worst_case_accuracy_language}
\resizebox{\textwidth}{!}{%
\begin{tabular}{lrrrrrrrr}
\hline
\textbf{Model} & \textbf{ALL} & \textbf{English} & \textbf{Arabic} & \textbf{Chinese} & \textbf{Hindi} & \textbf{Indonesian} & \textbf{Kazakh} & \textbf{Swahili} \\
\hline
\multicolumn{9}{c}{\textit{Closed-source Models}} \\
\hline
GPT-5.4 & 73.9 & 73.6 & 74.6 & 74.0 & 72.6 & 72.9 & 75.6 & 74.2 \\
Gemini-2.5-Pro & 62.5 & 64.8 & 61.6 & 59.8 & 63.7 & 62.1 & 61.9 & 63.4 \\
Gemini-2.5-Flash & 56.6 & 57.3 & 55.9 & 57.2 & 54.7 & 56.6 & 56.6 & 58.1 \\
\hline
\multicolumn{9}{l}{} \\
\multicolumn{9}{c}{\textit{Open-source Models}} \\
\hline
Qwen3.5-122B-A10B & 43.6 & 45.1 & 42.6 & 41.1 & 44.5 & 43.2 & 44.1 & 44.4 \\
Llama-4-Scout & 33.7 & 31.4 & 34.6 & 36.3 & 36.7 & 32.1 & 32.3 & 32.5 \\
InternVL3.5-14B & 18.4 & 22.4 & 22.4 & 18.6 & 19.0 & 21.5 & 14.9 & 9.8 \\
InternVL3.5-8B & 25.8 & 30.1 & 27.4 & 23.7 & 26.8 & 26.4 & 26.2 & 19.8 \\
InternVL3.5-4B & 17.5 & 19.0 & 18.4 & 19.1 & 17.6 & 18.6 & 18.3 & 11.3 \\
Qwen3-VL-8B-Instruct & 29.5 & 28.2 & 32.1 & 33.8 & 28.0 & 31.0 & 27.4 & 25.7 \\
Qwen3-VL-4B-Instruct & 3.3 & 3.7 & 3.5 & 3.3 & 3.0 & 4.0 & 2.7 & 3.2 \\
Molmo2-8B & 15.5 & 17.5 & 15.3 & 14.8 & 15.5 & 17.5 & 16.7 & 11.5 \\
Molmo2-4B & 10.9 & 14.3 & 10.9 & 13.5 & 11.7 & 11.8 & 8.5 & 5.3 \\
Gemma-3-12B-IT & 18.7 & 22.9 & 16.4 & 20.1 & 17.2 & 16.1 & 17.9 & 20.4 \\
Gemma-3-4B-IT & 7.2 & 8.3 & 6.9 & 6.4 & 6.8 & 7.6 & 7.5 & 6.7 \\
SEA-LION-v4-8B-VL & 33.3 & 33.4 & 33.8 & 32.1 & 33.9 & 36.4 & 34.5 & 29.1 \\
SEA-LION-v4-4B-VL & 10.0 & 11.6 & 9.2 & 9.9 & 9.6 & 11.4 & 9.8 & 8.9 \\
qwen-qwen3-5-27b & 51.2 & 57.0 & 49.2 & 48.2 & 49.1 & 53.9 & 47.8 & 53.0 \\
\hline
\end{tabular}%
}
\end{table}

\section{Robustness of the Step-Level Judge} \label{appendix:judge}
Using GPT-5.4 to judge its own reasoning steps raises a concern about self-preference bias, where an LLM judge favors responses generated by itself. This concern applies only to the step-level evaluation in Table~\ref{tab:reasoning_judge_subject_metrics}. All accuracy results (Tables~\ref{tab:stem_vision_english_avg_wc_by_subject}, \ref{tab:stem_vision_average_case_accuracy_language}, and \ref{tab:stem_vision_worst_case_accuracy_language}) are computed by matching the final answer against the ground truth with a 1\% tolerance, without any model involved. The judging task is also narrower than the open-ended comparisons where self-preference is usually observed. Since every template comes with a gold solution, the judge only decides whether a gold step is covered by the model output, or whether a model step is valid against the gold solution (Tables~\ref{tab:prompt_recall} and \ref{tab:prompt_precision}).

To verify that our step-level findings hold regardless of the judge, we re-evaluated the same GPT-5.4 outputs using Gemini 2.5 Pro as the judge with identical prompt templates. Table~\ref{tab:judge_comparison} compares the results. Scores under Gemini 2.5 Pro are slightly lower in most subjects, with the largest difference in Computer Science, where $\mathcal{F}$ drops by 4.6 points and $\mathcal{F}_{wst}$ by 9.1 points. Nevertheless, the gap between $\mathcal{F}$ and $\mathcal{F}_{wst}$ remains wide in every subject under both judges (13.0--20.8 points with GPT-5.4 and 15.5--25.0 points with Gemini 2.5 Pro), and Biology remains the most challenging subject. This indicates that the reasoning step fragility reported in Section~\ref{sec:main_result} holds under a judge from a different model family.

\begin{table}[h]
\centering
\caption{Step-level evaluation of GPT-5.4 outputs judged by GPT-5.4 and Gemini 2.5 Pro. $\Delta$ denotes the change when switching the judge from GPT-5.4 to Gemini 2.5 Pro.}
\label{tab:judge_comparison}
\begin{tabular}{l|rrr|rrr}
\hline
\textbf{Subject} & \multicolumn{3}{c|}{$\mathcal{F}$} & \multicolumn{3}{c}{$\mathcal{F}_{wst}$} \\
 & GPT-5.4 & Gemini-2.5-Pro & $\Delta$ & GPT-5.4 & Gemini-2.5-Pro & $\Delta$ \\
\hline
Mathematics      & 88.0 & 86.5 & -1.5 & 67.8 & 66.5 & -1.3 \\
Physics          & 92.2 & 92.4 & +0.2 & 79.2 & 76.7 & -2.5 \\
Chemistry        & 93.4 & 93.2 & -0.2 & 79.0 & 77.7 & -1.3 \\
Biology          & 81.1 & 80.0 & -1.1 & 60.3 & 55.0 & -5.3 \\
Computer Science & 89.4 & 84.8 & -4.6 & 75.1 & 66.0 & -9.1 \\
\hline
\end{tabular}
\end{table}

\end{document}